\documentclass[10pt,twocolumn,letterpaper]{article}

\usepackage[dvipsnames,table]{xcolor}
\usepackage{cvpr}              %

\usepackage{arydshln}
\usepackage{graphicx}
\usepackage{amsmath}
\usepackage{amssymb}
\usepackage{booktabs}
\usepackage{multirow}
\usepackage{rotating}
\usepackage[normalem]{ulem}
\useunder{\uline}{\ul}{}
\usepackage{epigraph} 
\usepackage{dsfont}
\usepackage{bm}
\usepackage{enumitem} %
\usepackage[accsupp]{axessibility}
\usepackage{float}  %

\usepackage{pifont}%
\newcommand{\cmark}{\ding{51}}%
\newcommand{\xmark}{\ding{55}}%

\definecolor{cvprblue}{rgb}{0.21,0.49,0.74}
\newcommand{\PAR}[1]{\noindent{\bf #1~}}

\newcommand{\method}{MATCHA\xspace}

\definecolor{cgreen}{RGB}{237, 252, 233}
\definecolor{cgrey}{RGB}{241, 241, 243}

\definecolor{cvprblue}{rgb}{0.21,0.49,0.74}
\usepackage[pagebackref,breaklinks,colorlinks,allcolors=cvprblue]{hyperref}

\def\paperID{6601} %
\def\confName{CVPR}
\def\confYear{2025}

\title{MATCHA\includegraphics[scale=0.1]{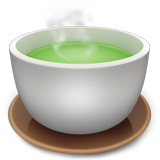}: Towards Matching Anything}

\author{Fei Xue$^1$\thanks{This work was done when Fei Xue was an intern at NVIDIA.} \,\,
Sven Elflein$^{2, 3, 4}$\,\,
Laura Leal-Taixé$^3$ \,\,
Qunjie Zhou$^3$
\\
$^1$University of Cambridge \quad
$^2$University of Toronto \quad
$^3$NVIDIA \quad
$^4$Vector Institute
}

\begin{document}
\maketitle

\begin{abstract}

Establishing correspondences across images is a fundamental challenge in computer vision, underpinning tasks like Structure-from-Motion, image editing, and point tracking. Traditional methods are often specialized for specific correspondence types, geometric, semantic, or temporal, whereas humans naturally identify alignments across these domains. Inspired by this flexibility, we propose \method, a unified feature model designed to ``rule them all'', establishing robust correspondences across diverse matching tasks. 
Building on insights that diffusion model features can encode multiple correspondence types, \method augments this capacity by dynamically fusing high-level semantic and low-level geometric features through an attention-based module, creating expressive, versatile, and robust features. 
Additionally, \method integrates object-level features from DINOv2 to further boost generalization, enabling a single feature capable of matching anything. 
Extensive experiments validate that \method consistently surpasses state-of-the-art methods across geometric, semantic, and temporal matching tasks, setting a new foundation for a unified approach for the fundamental correspondence problem in computer vision. To the best of our knowledge, \method is the first approach that is able to effectively tackle diverse matching tasks with a single unified feature.

\end{abstract}
    
\section{Introduction}
\label{sec:intro}
\begin{figure}[t!]
    \centering
    \includegraphics[trim=0.1cm 2.5cm 4.5cm 0cm,  width=\linewidth]{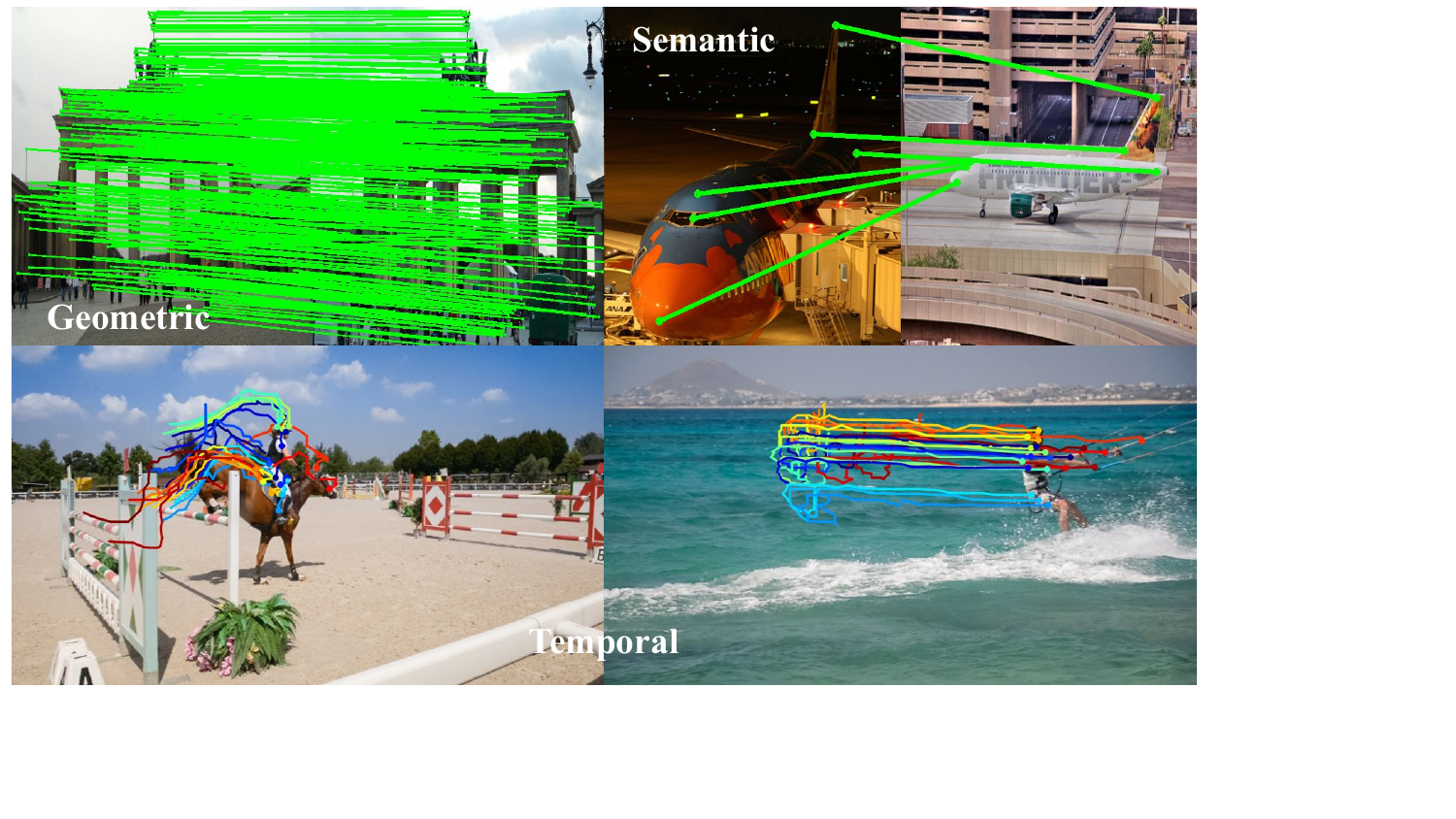}
    \caption{\textbf{\method for \textit{matching anything}.} We visualize  geometric, semantic and temporal correspondences established by \method, using a single feature descriptor.}
    \label{fig:correspondence_types}
    \vspace{-0.3cm}
\end{figure}

\textit{``In computer vision, there is only one problem: correspondence, correspondence, correspondence." --Takeo Kanade}
   \vspace{0.3cm}
   
Establishing correspondences between images is a fundamental problem in computer vision, 
integral to a variety of applications such as mapping and localization~\cite{schonberger2016sfm, mur2015orbslam}, image editing~\cite{ofri2023neural},  object pose estimation~\cite{xu2022rnnpose} and point tracking~\cite{doersch2022tapvid, harley2022particle}.
Correspondence is typically categorized by type: geometric~\cite{lowe2004sift, detone2018superpoint, revaud2019r2d2, sarlin2020superglue}, semantic~\cite{kim2017fcss, zhang2024geosddino,  zhang2024sddino} and temporal~\cite{tang2023dift, cho2024locotrack, karaev2023cotracker, sand2008particle} correspondences, as shown in \cref{fig:correspondence_types}.
Geometric correspondences identify 2D points in images of static scenes that represent the same physical 3D point, with challenges in diverse illumination and viewpoint variations. They are typically used to extract accurate geometric transformations between cameras \eg, for structure-from-motion applications.
Semantic correspondences connect similar object parts across distinct instances within a category, demanding high-level abstraction across different instances. 
Temporal correspondences, in contrast, match points of the same instance across video frames, require to handle both static and dynamic elements, occlusions, deformations and viewpoint changes stemming from complex motions.

Addressing these distinct challenges usually requires specialized models~\cite{tyszkiewicz2020disk, detone2018superpoint, potje2024xfeat, zhang2024sddino, li2024sd4match, luo2024dhf}.
However, humans can align points flexibly across different scenarios, \eg, across static scenes, dynamic objects of different instances under various viewpoints, prompting the question: \textit{Do we really need a separate feature for each type of correspondence problem?} 
DIFT~\cite{tang2023dift} offers a step toward this, revealing that correspondence patterns can emerge naturally from diffusion models~\cite{dhariwal2021amdiffusion, rombach2022stablediffusion}. 
However, DIFT still relies on distinct feature descriptors for different tasks, potentially limiting its utility when the matching type is unknown. More importantly, the unsupervised correspondences learned by DIFT fall short of fully supervised methods in matching accuracy (\cf \cref{sec:exp-semantic_matching} and \cref{sec:exp-geometric_matching}).

\begin{figure}[t!]
    \centering
    \includegraphics[trim=0.1cm 1.5cm 9.2cm 0.1cm, width=\linewidth]{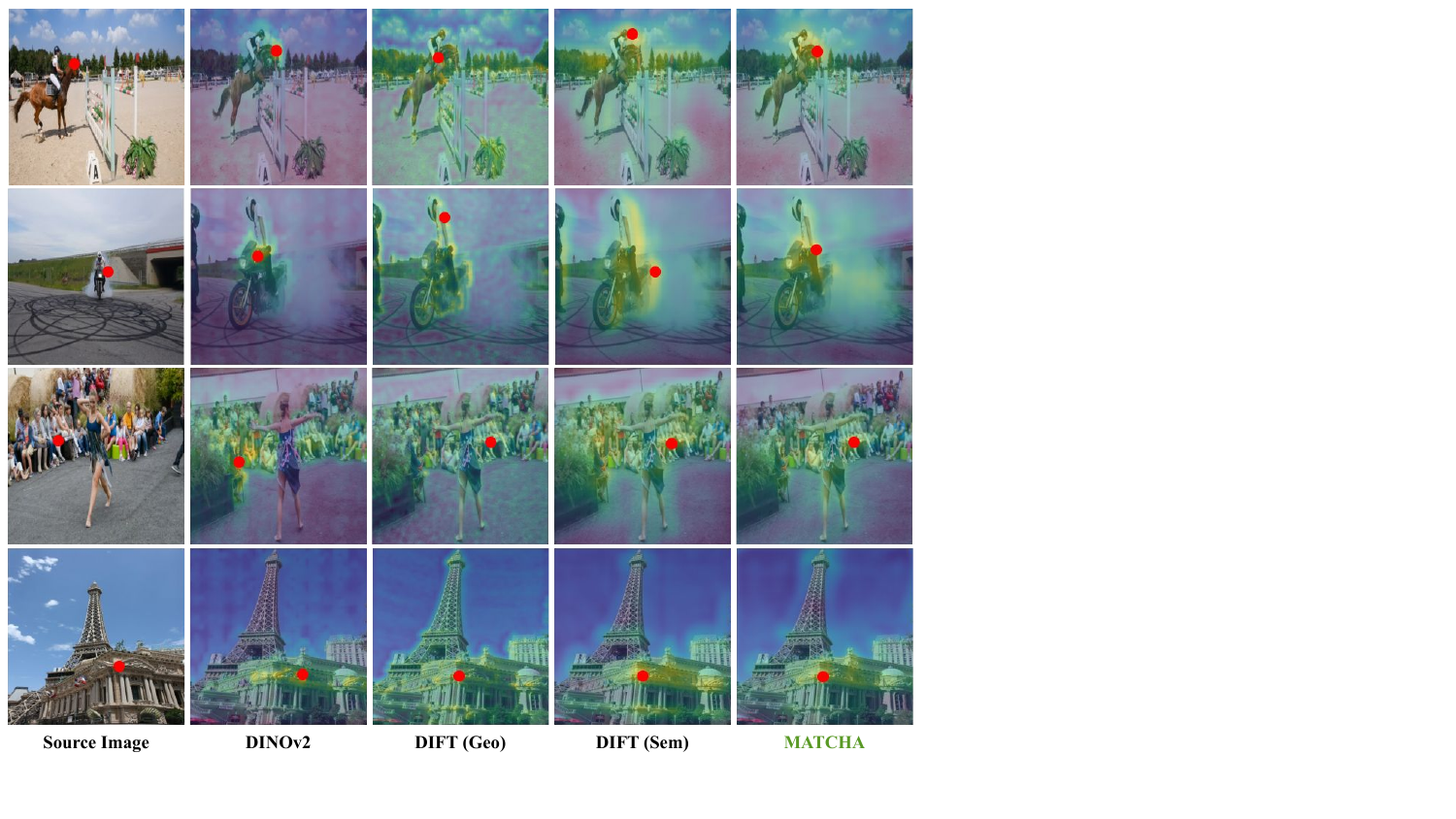}
    \caption{\textbf{Heatmap of features from DINOv2, DIFT, and MATCHA.} Given a query \textcolor{red}{point} from the source image (1st column), DINOv2 features give more accurate correspondences on single object (1st and 2nd row) but struggle when multiple instances of the same class (3rd row) or similar structures (4th row) exist. Both geometric and semantic features of DIFT perform reversely. By unifying knowledge in the three foundation features, \method produces more accurate and reliable correspondences.}
    \label{fig:heatmap}
    \vspace{-0.3cm}
\end{figure}

In this work, we introduce \method, a foundation feature model for \textit{matching anything}. Unlike DIFT, our approach learns a single feature descriptor for geometric, semantic, and temporal matching, incorporating explicit supervision while leveraging the rich knowledge of foundation models learned from large-scale data.
Our experiments confirm that combining foundation model knowledge with targeted supervision is key to accurate and generalized matching (\cf \cref{tab:model_ablation} and \cref{tab:towards_matching_anything}).
While adding correspondence-level supervision is straight-forward, there are only limited annotated datasets for supervision compared to the scale of data a foundation model is usually trained on, especially for semantic and temporal matching tasks where human annotations are required for real-world data.
Thus, the main challenge that we need to address is to find a proper way to inject accurate correspondence supervision from only a limited amount of annotated data, without destroying the rich information and generalization of features learned by foundation models.
To achieve this, we leverage an attention-based dynamic feature fusion module that learns to extract mutually supportive knowledge from two domains, \ie, semantic and geometric, to enhance themselves for improved matching performance.
Guided by correspondence-level supervision, our attention-based fusion enhances diffusion feature representations without losing generalization.
Supported by the fusion process, we are able to combine the enhanced diffusion features with the complementary semantic knowledge from DINOv2, which captures robust, single-object correspondences (as shown in Fig~\ref{fig:heatmap}). 
The result is a unified, high-quality feature that achieves strong matching performance across different tasks.

We summarize the contributions of this work as follows:
We (i) systematically analyze common feature models for matching, informing the design of \method, a novel feature model that learns to dynamically fuse geometric and semantic information to improve representational robustness without loss of generality.
\method demonstrates that (ii) static fusion of features can offset the limitations of individual descriptors, enabling a single feature to address a range of correspondence tasks effectively.
Comprehensive evaluations show (iii) \method surpasses state-of-the-art on most benchmarks, significantly outperforming unsupervised methods in semantic and geometric matching, highlighting the importance of correspondence supervision for precision.
For the first time, we show that (iv)  a single feature is able to achieve the new state-of-the-art across all three types of common correspondence problems.
(iiv) As a contribution to the community, we re-purpose the TAP-Vid point tracking benchmark~\cite{doersch2022tapvid} for temporal matching evaluation, establishing common feature baselines to support future research on unified feature learning for matching.

\section{Related Work}
\label{sec:related_work}

\PAR{Geometric Correspondence.}
Geometric matching refers to searching physically correspondent point pairs between two images captured in the same scene. 
Geometric correspondences are commonly established by detecting, describing, and matching local features. An abundant of local feature detection and description methods have been developed starting from hand-crafted local features such as SURF~\cite{bay2008surf} and SIFT~\cite{lowe2004sift} and then evolving towards learned ones~\cite{yi2016lift, detone2018superpoint, dusmanu2019d2net, revaud2019r2d2, xue2023sfd2, tyszkiewicz2020disk, luo2019contextdesc, tian2019sosnet}.
Benefiting from massive training data, the learned features show better discriminative ability to viewpoint and illumination changes than handcrafted ones. However, as most of these learned features are trained with purely geometric ground-truth correspondences mainly from static objects, despite their promising accuracy on geometric matching, they have poor performance especially on semantic matching (\cf \cref{sec:exp-semantic_matching} and \cref{sec:exp-geometric_matching}). 
Geometric correspondences can be obtained with nearest neighbor matching~\cite{muja2014nnm} based on descriptor distance. Although more powerful learned sparse~\cite{sarlin2020superglue, lindenberger2023lightglue, jiang2024omniglue} and dense~\cite{rocco2018ncnet, zhou2021patch2pix, germain2020s2dnet, sun2021loftr, edstedt2023dkm, chen2022aspanformer, edstedt2024roma} matchers are proposed, in this paper, we focus mainly on the feature itself and use nearest neighbor matching to find correspondences.

\PAR{Semantic Correspondence.}
Semantic matching aims to match points with similar semantic meaning across different instances of the same category, \eg,  matching the eyes of a cat in one image to another cat in the other image. 
Semantic matching methods focus on extracting feature descriptors~\cite{choy2016ucn, kim2017fcss} to capture semantic information.
Recent works~\cite{zhang2024geosddino, zhang2024sddino, hedlin2024usc, luo2024dhf, li2024sd4match, tang2023dift} leverage features extracted from foundation models~\cite{caron2021dino, oquab2023dinov2, radford2021clip, dhariwal2021amdiffusion, rombach2022stablediffusion} due to their rich semantic knowledge which is hard to learn from a limited amount of annotated semantic matching training data. These methods, \eg, DIFT~\cite{tang2023dift} and SD+DINO~\cite{zhang2024sddino} use the foundation model features directly for semantic matching. However, their performance is not comparable to those finetuned with supervision, \eg, DHF~\cite{luo2024dhf} and SD4Match~\cite{li2024sd4match}. Some works also build semantic matchers for matching from the perspective of customized matching functions~\cite{liu2010siftflow, kim2013dsp,lee2019sfnet}, correspondence networks~\cite{rocco2017geometric, rocco2018geometric} or semantic flow~\cite{kim2019semantic, rocco2018ncnet, lee2019sfnet, truong2020glu,cho2022cats++, huang2022scorrsan}. These methods require paired images as input rather than single images.

\PAR{Temporal Correspondence.}
Temporal matching targets at establishing correspondences of the same object across video frames. 
It generalizes the geometric matching task from static scenes to general natural scenes that contain both static and dynamic content.
Recently, temporal correspondence has been largely investigated in its downstream application task, \ie, tracking any point (TAP)~\cite{doersch2022tapvid}.
The point tracking work~\cite{sand2008particle, harley2022particle, doersch2022tapvid, karaev2023cotracker, cho2024locotrack} focuses on occlusion handling and exploring temporal priors, \eg, long-term consistency, motion constraints, as well as leveraging 3D reconstruction~\cite{Xiao2024spatialtracker, wang2023omnimotion, luiten2023dynamic, som2024shapeofmotion, seidenschwarz2024dynomo}.
Compared to these works, we are interested in the general problem of establishing pair-wise correspondences of any two frames from a video without leveraging any temporal constraints.

\PAR{Vision Foundation Model.}
Modern vision foundation models, \eg, DINO~\cite{caron2021dino, oquab2023dinov2}, CLIP~\cite{radford2021clip, ilharco2021openclip}, and diffusion models~\cite{rombach2022stablediffusion, dhariwal2021amdiffusion}, exhibit strong generalization performance across a variety of tasks or domains.
Excitingly, their features show promising accuracy for both geometric~\cite{edstedt2024roma, jiang2024omniglue} and semantic~\cite{luo2024dhf, rocco2017geometric, li2024sd4match, hedlin2024usc} correspondences, or even directly delivering emergent correspondences without an explicit supervision~\cite{zhang2024sddino, tang2023dift}. DIFT~\cite{tang2023dift} demonstrates that rich semantic and geometric features have been learned by image diffusion models and can be utilized to directly establish semantic, geometric and temporal correspondences without further supervision. SD+DINO~\cite{zhang2024sddino} reveals that features from different foundation models have different properties and demonstrates that the combination of SD feature and DINO feature gives better semantic accuracy than either of them. Inspired by these two works, we leverage SD model and DINOv2 as our backbones to provide raw features. However, essentially different with these two works, we focus on how to obtain a single feature for all three types of matching by involving the supervision signals.

\begin{figure*}[t!]
    \centering
    \includegraphics[trim=0.5cm 0.5cm 1.5cm 0cm, width=0.98\linewidth]{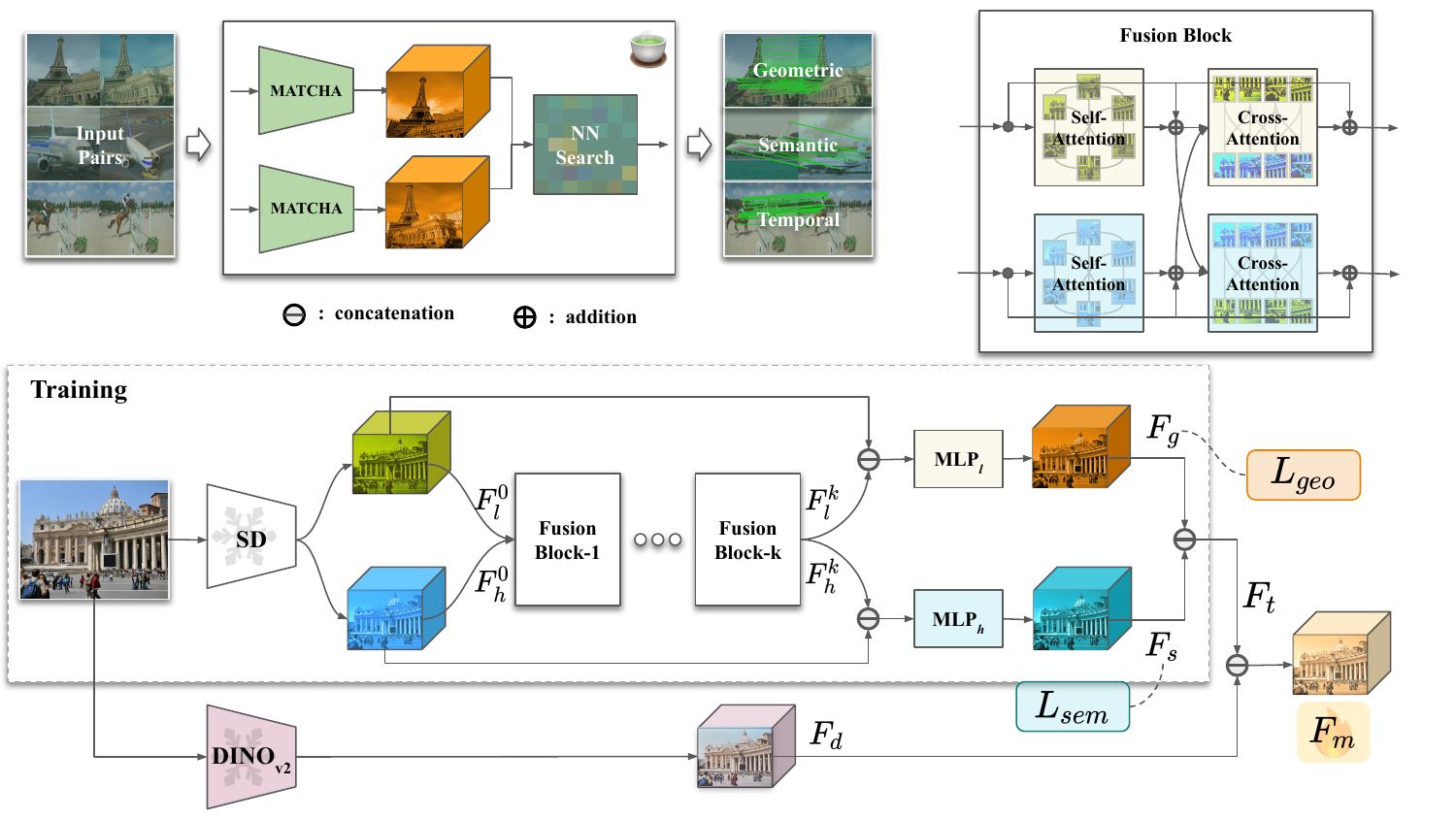}
    \caption{\textbf{Architecture of \method.} Given an RGB image, \method produces a single feature for geometric, semantic and temporal matching with nearest neighbor searching. \method is built on top of stable diffusion (SD) models~\cite{rombach2022stablediffusion} and DINOv2~\cite{oquab2023dinov2}. Specifically, original geometric and semantic features extracted from SD are first fused dynamically with a transformer~\cite{vaswani2017attention} consists of self and cross attention blocks. In this dynamic fusion process, both geometric and semantic features are augmented with each other which are supervised with corresponding ground-truth signals in the training process. Then, augmented geometric and semantics features along with DINOv2 feature are unified statically via concatenations into a single feature for \textit{matching anything}. 
    }
    \label{fig:architecture}
\end{figure*}

\section{\method}
\label{sec:method}

In this section, we present \method, a novel feature model that unifies knowledge from multiple foundation models\cite{rombach2022stablediffusion, dhariwal2021amdiffusion, caron2021dino, oquab2023dinov2} and enhances features for accurate correspondences through precise supervision, enabling a single feature descriptor for correspondence problems across different domains, reaching the state-of-the-art performance.

\subsection{Preliminary}
\label{sec:method:preliminary}
Our method is inspired by previous work DIFT that extracts features from a diffusion model for unsupervised matching. We also build on top of DINOv2~\cite{oquab2023dinov2}, a powerful self-supervised foundation model.

\PAR{DIFT~\cite{tang2023dift}.} 
Recent work, DIFT, demonstrates that diffusion models trained for pixel-wise image generation implicitly learn correspondences. By extracting features from specific layers and timestamps, DIFT~\cite{tang2023dift} identifies effective feature descriptors for geometric, semantic, and temporal matching tasks. Given an RGB image \(I \in \mathbb{R}^{H \times W \times 3}\), DIFT extracts a semantic descriptor \(F_h \in \mathbb{R}^{H/16 \times W/16 \times 1280}\) and a geometric descriptor \(F_l \in \mathbb{R}^{H/8 \times W/8 \times 640}\) from a pre-trained stable diffusion model~\cite{rombach2022high}. While the semantic descriptor \(F_h\) is used for semantic matching and \(F_l\) for geometric and matching, DIFT requires manual selection of descriptors per task, which limits flexibility and generalization.
Our approach eliminates the need for task-specific descriptors, achieving greater accuracy across tasks while maintaining a single, unified descriptor.

\PAR{DINOv2~\cite{oquab2023dinov2}.} 
DINOv2 is a self-supervised model trained on millions of images for object- and patch-level discrimination, which enables its features to capture rich semantic information for establishing object-level correspondences, as shown in recent work~\cite{zhang2024sddino}. In our experiments, DINOv2 also demonstrates robust handling of extreme viewpoint changes and scale variations for individual objects, excelling in temporal matching tasks (\cf \cref{sec:exp-temporal_matching}). 
While DINOv2 and DIFT (\(F_h\)) both provide semantic descriptors, our results show complementary strengths between the two that enhance general matching capability (\cf \cref{tab:model_ablation}). However, DINOv2’s lack of spatial detail limits its geometric matching performance. Our approach integrates knowledge from stable diffusion and DINOv2, unifying them into a powerful, single representation for matching across diverse tasks. We denote the feature extracted from DINOv2 as \(F_d \in \mathbb{R}^{H/14 \times W/14 \times 1024}\).

\subsection{Architecture}
\label{sec:method:architecture}
\label{subsec:achitecture}

\PAR{Overview.} 
As shown in \cref{fig:architecture}, given an RGB image $I$ as input, \method outputs a single feature descriptor $F_m\in R^{H/8\times W/8 \times D_m}$ ($D_m$ is its channel size) for correspondence matching, including geometric, semantic and temporal matching tasks.
(i) First, we build on top of the foundation feature models, DIFT and DINOv2, by obtaining two semantic feature descriptors $F_h$ and $F_d$ and a geometric descriptor $F_l$ (\cf \cref{sec:method:preliminary}).
 (ii) We next enhance the DIFT geometric and semantic features $F_l$ and $F_h$ by learning to extract supportive information from the other domain's descriptor. 
Such dynamic fusion is learned via correspondence-level joint supervision on semantic and geometric matching. We show in our later ablations that such a learned dynamic fusion is critical for a successful and balanced merging stage where each descriptor can build on top of each other.
(iii) Finally, we directly merge the two enhance features and the DINOv2 semantic feature $F_d$ into a single unified feature $F_m$.
We describe the detail of each step in the followings.

\PAR{Dynamic feature fusion.} 
We adopt the transformer~\cite{vaswani2017attention} with self and cross attention mechanism for fusion. This strategy allows our model to dynamically gather complementary information from the geometric and semantic descriptors and supervise them jointly in the training process. We patchify both features $F_h$ and $F_l$ with a patch size of $p$ and project their feature dimension to a common feature dimension $D_h$ with a linear layer, which produces the input semantic feature $F^{0}_h\in R^{N \times D_h}$ and geometric feature $F^{0}_l\in R^{N \times D_h}$ for the fusion stage, where $N=\frac{H}{p*8}\times \frac{W}{p*8}$ is the number of patchified features. The fusion module consists of $k$ self- and cross-attention blocks.
For $i$-th block with $i\in\{1,...,k\}$, the updating process is as follows:
\begin{align}
    F^{i}_{hs} &= F^{i-1}_h + \mathtt{self}_{h}^{i}(F^{i-1}_h),\\
    F^{i}_{ls} &= F^{i-1}_l + \mathtt{self}_{l}^{i}(F^{i-1}_l), \\ 
    F^{i}_h &= F^{i-1}_h + \mathtt{cross}_{h}^{i}(F^{i}_{hs},F^{i}_{ls}), \\    
    F^{i}_l &= F^{i-1}_l + \mathtt{cross}_{l}^{i}(F^{i}_{ls}, F^{i}_{hs}), 
    \label{eq:fusion}
\end{align}
where $\mathtt{self}_{h}^{i}$ and $\mathtt{self}_{l}^{i}$ are $i$-th self-attention blocks for $F_h$ and $F_l$, respectively. $ \mathtt{cross}_{h}^{i}$ and $ \mathtt{cross}_{l}^{i}$ are $i$-th  cross-attention blocks for $F_h$ and $F_l$, respectively. We use the same multi-head attention architecture for each feature branch with non-sharing parameters.
Finally, we concatenate the original input features and the fused features along the channel dimension and feed them into a two-layer MLP to output the final semantic feature $F_s$ and geometric feature $F_g$, defined as:
\begin{align}
    F_s =  \mathtt{MLP}_h([F^{0}_h||F^{k}_h]), F_g =  \mathtt{MLP}_l([F^{0}_l||F^{k}_l]),
\end{align}
where $[.||.]$ denotes channel-wise concatenation. $F_g$ and $F_s$ are augmented geometric and descriptors and can be used directly for geometric and semantic matching, respectively.

\PAR{Feature Merging.} 
With the previous preparation of the fusion, we are able to smoothly merge the three features to unify their knowledge.
We start by concatenating the enhanced semantic and geometric features, $F_s$ and $F_g$, to form $F_t$, which effectively captures both semantic and geometric information within the image.
As shown in \cref{tab:model_ablation}, this explicit merging, built upon the dynamic fusion process, results in a single feature that significantly outperforms the direct merging of raw DIFT features without fusion enhancement. This demonstrates its superior ability to handle both semantic and geometric information simultaneously.
To further boost its matching ability, we equip $F_t$ with the strong semantic cues of DINO-v2 in $F_d$ by another concatenation to obtain the final unified matching feature $F_m$. 
Specifically, the two concatenations are defined by:
\begin{align}
    F_t = (F_g||F_s(...,::d_s)), F_m = (F_t||F_d(...,::d_t)),
\end{align}
where $d_s=\frac{D_s}{D_g}$ and $d_t=\frac{D_d}{D_t}$ are strides adopted to downsample $F_s$ and $F_d$ along the channel dimension.

\subsection{Supervision}
\label{sec:method:supervision}
Instead of providing supervision on the final unified feature, we choose to only provide supervision signals to the dynamic fusion enhancement.
Ideally, we want to introduce precise signals on each of the tasks directly to our unified feature, which usually requires large-scale annotated data for balanced training across different tasks. 
However, it is highly expensive to obtain large-scale and accurate correspondence annotations, especially for semantic matching and temporal matching. 
Therefore, with the limited amount of supervision, we choose to customize the DIFT feature for semantic matching, and support the general semantic understanding from DINOv2 descriptor without further tuning it.
Specifically, we apply semantic matching supervision to $F_s$ using CLIP contrastive loss~\cite{radford2021clip} combined with a dense semantic flow loss~\cite{lee2019sfnet} and geometric matching supervision to $F_g$ using the dual softmax loss function~\cite{potje2024xfeat}. We provide more information about our supervision losses and training details in the supplementary material.

\begin{table}[t]
\centering
\resizebox{8cm}{!} {
\begin{tabular}{lcccc}
\toprule
    &  SM. &SPair-71k~\cite{min2019spair}&PF-Pascal~\cite{ham2017pfpascal}&PF-Willow~\cite{ham2016pfwillow}\\
    Method & Sup.  & PCK$_{@0.01/0.05/0.1}(\uparrow)$ & \multicolumn{2}{c}{PCK$_{@0.05/0.1/0.15}(\uparrow)$} \\ 
\midrule
DINOv2~\cite{oquab2023dinov2}  & \xmark &6.3 / 38.4 / 53.9 &63.0 / 79.2 / 85.1& 43.8  /  75.4 / 86.1 \\
$^{\star}$DIFT~\cite{tang2023dift}  & \xmark &7.2 / 39.7 / 52.9 & 66.0 / 81.1 / 87.2 & 58.1 / 81.2 / -  \\
DIFT & \xmark                 & 3.1 / 37.9 / 54.3 & 58.7 / 81.8 / 87.8 & 55.7 / 85.1 / 92.9\\
USC~\cite{hedlin2024usc}       & \xmark &- / 28.9 / 45.4 & - &53.0 / 84.3 /  -    \\
SD+DINO~\cite{zhang2024sddino} & \xmark &7.9 / 44.7 / 59.9 &71.5 / 85.8 / 90.6 &  -   \\
$^{\dagger}$GeoASM~\cite{zhang2024geosddino} & \xmark &9.9 / 49.1 / 65.4 & 74.0 / 86.2 / 90.7 & - \\

\midrule
DHF~\cite{luo2024dhf}  & \cmark &8.7 / 50.2 / 64.9& 78.0 / 90.4 / 94.1 &  -  \\
\textbf{*}\textcolor{red}{SCorrSAN}~\cite{huang2022scorrsan} & \cmark &3.6 / 36.3 / 55.3 &81.5 / 93.3 / 96.6 & 54.1 /  80.0  / 89.8    \\
\textbf{*}\textcolor{red}{CATs++}~\cite{cho2022cats++}  & \cmark &4.3 / 40.7 / 59.8& \underline{84.9} / 93.8 / 96.8 & 56.7 / - / 81.2    \\
\textbf{*}SD4Match~\cite{li2024sd4match} & \cmark & -  / 59.5 / 75.5 & 84.4 / \underline{95.2} / \underline{97.5} & 56.7 / 80.9 / 91.6 \\

\textbf{*}SD+DINO ~\cite{zhang2024sddino}  & \cmark &9.6 / 57.7 / 74.6& 80.9 / 93.6 / 96.9&  -    \\
\textbf{*}$^{\dagger}$GeoASM ~\cite{zhang2024geosddino} & \cmark & \textbf{22.0 / 75.3 / 85.6}& \textbf{85.9 / 95.7 / 98.0} &-  \\

\midrule 
\textbf{\method-Light} & \cmark & 10.4 / 65.5 / 78.9 & 82.3 / 93.5 / 96.6 & \underline{69.0 / 90.1 / 96.2}\\
\textbf{\method}  &  \cmark & \underline{12.2 / 67.1 / 79.6} & 79.5 / 93.0 / 96.8 & \textbf{70.2 / 91.3 / 97.0}\\
\bottomrule
\end{tabular}
}
\caption{\textbf{Evaluation on Semantic Matching.} We report PCK under different thresholds. \textbf{*} denotes methods with dataset-specific models and $\dagger$ denotes semantic masks being required. \textcolor{red}{Red} indicates methods using image pairs as inputs. Both results of DIFT from its original paper~\cite{tang2023dift} ($^{\star}$DIFT) and our implementation (DIFT) are included.}
\label{tab:semantic_matching}
\vspace{-0.3cm}

\end{table}

\section{Experiments}
\label{sec:experiments}

We evaluate \method on three matching tasks. We also test a variant of our method, denoted as \method-Light, which evaluates the individual performance of $F_s$, $F_g$ and $F_t$ (\cf \cref{sec:method:architecture}) on semantic, geometric, and temporal matching tasks. This model is lighter due to no fusion from DINOv2, and follows DIFT by tackle different matching tasks using different descriptors. More experiments are provided in the supplementary material.

\subsection{Semantic Matching}
\label{sec:exp-semantic_matching}

\PAR{Datasets.} Following~\cite{zhang2024geosddino, li2024sd4match, cho2022cats++}, we use three widely used datasets. SPair-71k~\cite{min2019spair} contains 12,234 testing pairs split from 70,958 annotated pairs across 18 classes, with diverse scenes and significant viewpoint and scale variation. 
PF-PASCAL~\cite{ham2017pfpascal} includes 299 testing pairs split from 3547 annotated pairs with similar viewpoints and instance pose. 
PF-WILLOW~\cite{ham2016pfwillow} contains 900 testing pairs across 4 categories and is used to verify the generalization capability. We evaluate all datasets at an image resolution of $512 \times 512$.

\PAR{Baselines.} The baseline methods include those without supervision, \eg, DIFT~\cite{tang2023dift}, USC~\cite{hedlin2024usc}, DINOv2~\cite{oquab2023dinov2} as well as those supervised with GT semantic correspondences, \eg, SD4Match~\cite{li2024sd4match} and DHF~\cite{luo2024dhf}. 
We also show results of SD+DINO~\cite{zhang2024sddino} and GeoASM~\cite{zhang2024geosddino} which provide models with and without supervision. Besides, numbers of two semantic matchers SCorrSAN~\cite{huang2022scorrsan} and CATs++~\cite{cho2022cats++}  are also included as a reference.

\PAR{Metrics.} %
We adopt the standard metric of Percentage of Correct Keypoints (PCK) under different thresholds ($0.01/0.05/0.1$ for SPair and $0.05/0.1/0.15$ for others).

\PAR{Results.} 
As shown in \cref{tab:semantic_matching}, both \method and \method-Light surpass all other semantic features except for GeoASM~\cite{zhang2024geosddino} which requires dataset-specific trained models for evaluation and applies task-specific augmentation on top of its baseline SD+DINO~\cite{zhang2024sddino}. Such test-time augmentation requires masks of the dominant object to flip test images and is not applicable for geometric matching and temporal matching.
In contrast, we pursue general improvement in feature representation to better handle matching across various situations using a single feature model (\cf \cref{tab:towards_matching_anything}).
We show that our models stand out on PF-Willow, indicating strong generalization capability.

\begin{figure}[t]
\centering
    \includegraphics[width=\linewidth]{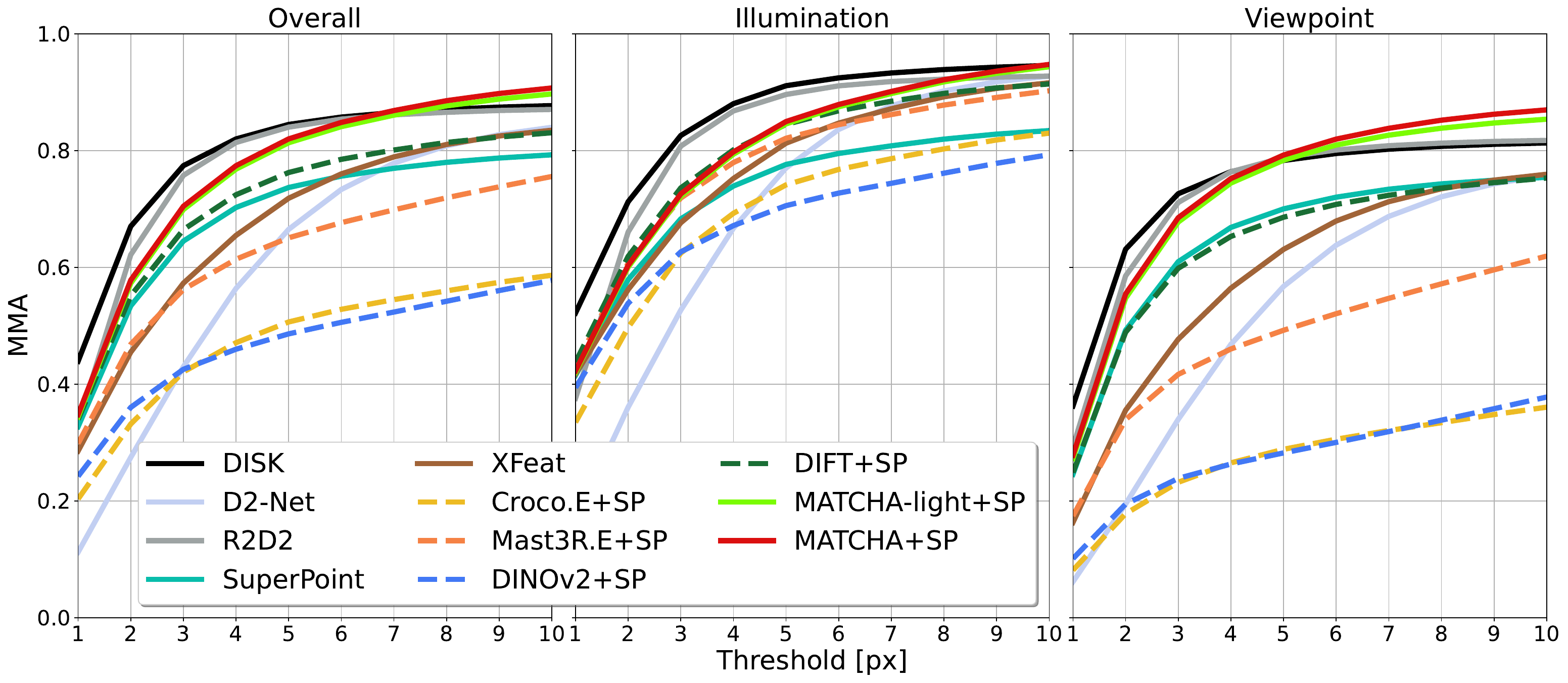}
    \caption{\textbf{Geometric Matching on HPatches}.  We report Mean Matching Accuracy (MMA) at error thresholds ranging from 1-10 pixel. Concrete and dash lines denote methods with and without supervision, respectively.}
\label{fig:hpatches_mma}
\vspace{-0.3cm}
\end{figure}

\subsection{Geometric Matching}
\label{sec:exp-geometric_matching}

\PAR{Datasets.} Following prior works~\cite{revaud2019r2d2, dusmanu2019d2net, tyszkiewicz2020disk, potje2024xfeat}, HPatches~\cite{balntas2017hpatches} is used to test feature matching performance.  We also utilize testing splits~\cite{sun2021loftr} of ScanNet~\cite{dai2017scannet} and  Megadepth~\cite{li2018megadepth} to evaluate the relative pose estimation. We further create randomly selected 1500 pairs of images with large viewpoint and appearance changes from the database of Aachen (Day\&Night) v1.0~\cite{sattler2018benchmarking} to validate the generalization ability. These four datasets cover geometric correspondences induced by homography and perspective transformations under indoor, outdoor and planar scenes with moderate to strong viewpoint and illumination changes.

\PAR{Baselines.} Our baselines include local features (\eg, SuperPoint (SP)~\cite{detone2018superpoint}, DISK~\cite{tyszkiewicz2020disk}, \etc),  foundation models (\eg, DINOv2~\cite{oquab2023dinov2}, DIFT~\cite{tang2023dift}) as well as encoders of recent popular geometric foundation models (\eg, Croco.E~\cite{weinzaepfel2022croco} and MASt3R.E~\cite{leroy2024mast3r}). Following DIFT~\cite{tang2023dift}, we use SP to provide keypoints for DINOv2, Croco.E, MASt3R.E and our method \method and \method-Light. We run all methods in the same setting on the original image resolution. We compute matches using nearest neighbour matching with mutual check and estimate relative pose using Poselib~\cite{larsson2020poselib} with LO-RANSAC~\cite{chum2003loransac} as XFeat~\cite{potje2024xfeat}.

\PAR{Metrics.} Following~\cite{dusmanu2019d2net, zhou2021patch2pix}, we report the mean matching accuracy (MMA) under 1-10 pixel error thresholds on HPatches and report the area under the curve (AUC) of poses accuracy at error thresholds of $5/10/20$ degrees for relative pose estimation.

\PAR{Feature matching results.} 
As shown in \cref{fig:hpatches_mma}, both our models have rather close performance on planar scenes, achieving overall the best matching accuracy at bigger thresholds, \eg, above 7px errors. 
At smaller thresholds, we are only less accurate than DISK and R2D2 which benefit from feature maps at the original image resolution. Note that all other methods including our models use downscaled feature maps ($8\times$ downsampling), but our models give the best accuracy among them. 

We observe that supervised methods, \eg, DISK, are much better than methods without supervision, \eg, DIFT, at handling viewpoint variations. This strongly suggests that accurate geometric matching against viewpoints is rather hard to learn from large data without precise correspondence supervisions.

\begin{table}[t]
\centering
\resizebox{8cm}{!} {
\begin{tabular}{lc c c c}
    \toprule
         \multirow{2}{*}{Method} & GM & MegaDepth~\cite{li2018megadepth} & ScanNet~\cite{dai2017scannet} & Aachen~\cite{sattler2018benchmarking} \\    
     & Sup. &  \multicolumn{3}{c}{AUC$_{@5/10/20}(\uparrow)$} \\

    \midrule 
        Croco.E~\cite{weinzaepfel2022croco} + SP & \xmark & 8.0 / 14.7 / 24.2 & 1.8 / 4.2 / 8.4  & 11.4 / 18.2 / 26.3 \\
        DINOv2~\cite{oquab2023dinov2} + SP & \xmark & 24.6 / 37.4 / 50.9 & 2.3 / 5.9 / 12.3 &  17.2 / 26.1 / 36.4\\  

        DIFT~\cite{tang2023dift} + SP &  \xmark & 49.7 / 62.8 / 72.8 & 9.3 / 18.7 / 29.4 & 43.7 / 53.1 / 61.3\\    
    \midrule
        SP~\cite{detone2018superpoint} & \cmark & 47.2 / 60.0 / 69.9 & 6.8 / 14.9 / 24.7 &  41.6 / 50.2 / 58.1\\
        XFeat~\cite{potje2024xfeat} & \cmark & 45.4 / 58.9 / 69.3 & 12.3 / 25.9 / 40.6 & 36.1 / 45.9 / 55.1 \\
	DISK~\cite{tyszkiewicz2020disk} & \cmark & 55.4 / 67.7 / 76.7 & 6.8 / 14.9 / 24.7 & 48.9 / 57.5 / 64.6 \\
	R2D2~\cite{revaud2019r2d2} & \cmark & 39.6 / 54.3 / 66.2 & 5.4 / 11.3 / 19.3 & 27.6 / 36.4 / 44.1 \\
        D2Net~\cite{dusmanu2019d2net} & \cmark & 32.5 / 47.7 / 61.4 & 10.6 / 22.9 / 37.3 & 30.3 / 41.8 / 52.5 \\  
        MASt3R.E~\cite{leroy2024mast3r} + SP & \cmark & 37.8 / 51.6 / 63.6 & 7.4 / 16.8 / 28.5 & 31.2 / 41.3 / 51.3\\ 
    \midrule 
    \textbf{\method-Light} + SP & \cmark & \textbf{57.1} / \textbf{70.9} / \textbf{81.2} & \textbf{13.0} / \textbf{26.6} / \textbf{41.8} & \underline{51.4 / 60.1 / 67.1}\\
    \textbf{\method} + SP & \cmark & \underline{55.8 / 69.3 / 80.0} & \underline{12.7 / 26.1 / 40.8} & \textbf{51.7 / 61.0 / 68.5}\\

	\bottomrule	
\end{tabular}
}
\caption{\textbf{Evaluation on Relative Pose Estimation.} We report the AUC values at error thresholds of $5^\circ/10^\circ/20^\circ$ on all datasets.
}
\vspace{-0.15cm}
\label{tab:relapose_est}

\end{table}

\PAR{Relative pose estimation results.}
As shown in \cref{tab:relapose_est}, our models achieve the best performance on both indoor and outdoor datasets. 
While DIFT is the most superior unsupervised method, we are able to drastically increase its AUC score by 6.1-8.4 point and 5.8-7.1 point on outdoor MegaDepth and Aachen, and by 3.7-11.4 point on indoor ScanNet. 
Additionally, even if trained with a huge number of 3D correspondences, the encoder of MASt3R is significantly less accurate than other feature models that are supervised with much less yet explicit geometric correspondences, \eg, DISK, SP and both of our models.
Those observations further validate the importance of a precise supervision direct on descriptors for robust and accurate geometric correspondences.

\begin{figure}[t!]
    \centering
    \includegraphics[width=\linewidth]{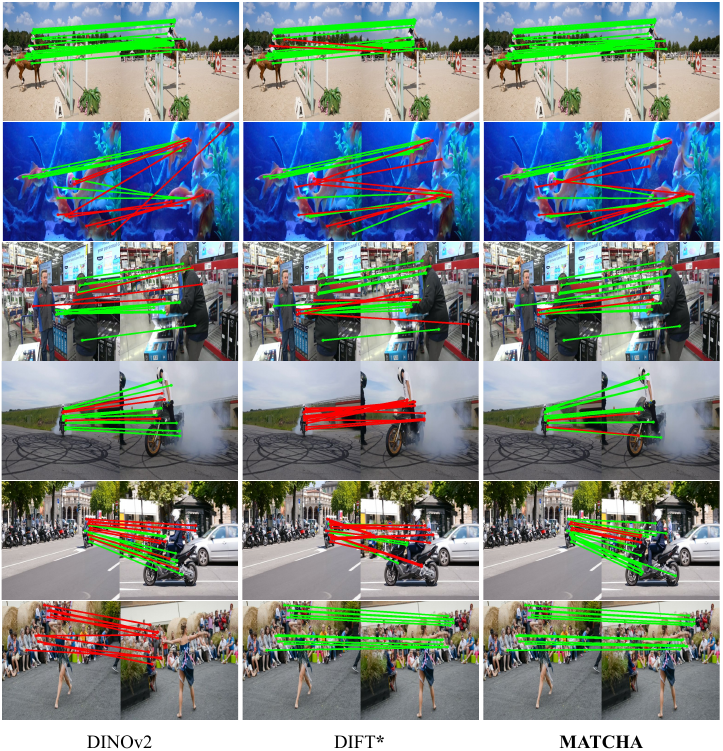}
    \caption{\textbf{Visualization of temporal matches on TapVID-Davis~\cite{doersch2022tapvid}.} 
    Here we visualize several challenging cases for exstablishing temporal correspondences, where \method generally achieves the best performance in handling extreme scale and viewpoint changes,  as well as scenes with multiple similar instances. (DIFT* is the adapted DIFT where we use its concatenated semantic and geometric feature for temporal matching for better performance. )
    }
    \label{fig:temporal_matching}
    \vspace{-0.3cm}
\end{figure}

\subsection{Zero-shot Temporal Matching}
\label{sec:exp-temporal_matching}

We further evaluate the zero-shot performance of our models on the challenging temporal matching task.%

\PAR{Datasets.} We re-purpose the existing TAPVid dataset~\cite{doersch2022tapvid} to benchmark feature models for temporal matching. TAPVid dataset consists of 30 highly varying real-world video sequences with unknown camera poses, among which some contain highly dynamic objects and extreme camera motions. We perform matching between the first frame and all following frames in each sequence to test the ability of features on handling temporal challenges.

\PAR{Baselines.} We compare our models to previous state-of-the-art geometric (\eg, DISK~\cite{tyszkiewicz2020disk}) and semantic (\eg, DIFT~\cite{tang2023dift}) matching baselines. Rather than using DIFT original feature for temporal matching as in their paper, we instead follow \method-Light to use its concatenated geometric and semantic feature, which leads to better performance.
We further consider a hybrid version of DIFT, DIFT.Uni+DINOv2, which combines geometric and semantic DIFT features as well as DINOv2 descriptors as in \method and can be considered as an unsupervised version of \method. 

\PAR{Metrics.} As TAPVid provides sparse query points for images, we report the same PCK metric at thresholds of $0.05/0.1/0.15$ as in semantic matching (\cf \cref{sec:exp-semantic_matching}).

\PAR{Results.} As shown in \cref{tab:towards_matching_anything}, among supervised methods, the geometric-matching-only models are generally better on temporal matching than the semantic-matching-only models.
However, among unsupervised methods, DINOv2 despite of its poor geometric matching performance (\cref{sec:exp-geometric_matching}) and moderate semantic matching capability (\cref{sec:exp-semantic_matching}), achieves surprisingly superior temporal matching ability.
While those two observations seem contradictory, our hypothesis is that DINOv2, benefiting from its large-scale learning on single object-centric data, is able to well handle large viewpoint and scale changes especially when there is a single dominant object in the scene.
However, it is poor at handling repetitive structures, and therefore it fails to achieve good geometric matching as well as temporal matching when many similar instances exist. We provide a visual example in \cref{fig:temporal_matching} that supports our hypothesis.

Furthermore, \method and DIFT.Uni+DINOv2 standing on top of DINOv2 are significantly better than the other baseline models, which infers that part of the semantic knowledge required for tackling temporal matching is uniquely supported by DINOv2.
Finally, \method outperforms DIFT.Uni+DINOv2, indicating that the accurate correspondence supervision signals from semantic and geometric matching provide additional help to improve temporal matching accuracy as well.

\begin{table}[t!]
\centering
\resizebox{8cm}{!} {
\begin{tabular}{lcccccc}
    \toprule
                       &  with  & Feat. & Corres.&Desc. &  Aachen & PF-Willow \\
     Baseline &  DINOv2 & Fusion & Sup   &Type & AUC$_{@5/10/20}(\uparrow)$ & PCK$_{@0.05/0.1/0.15}(\uparrow)$ \\
    \midrule                 

        DIFT    & \xmark & \xmark & \xmark & SM & \cellcolor{cgrey}{25.6 / 35.6 / 46.3} & \cellcolor{cgrey}{55.7 / 85.1 / 92.9} \\
        DIFT.S & \xmark & \xmark & \cmark & SM & \cellcolor{cgrey}{11.5 / 18.6 / 27.7 } &  \cellcolor{cgreen}{63.6 / 88.4 / 95.7 } \\    
    \textbf{\method-Light}  & \xmark & \cmark & \cmark & SM & \cellcolor{cgrey}{21.9 / 31.4 / 41.3}&  \cellcolor{cgreen}{69.0 / 90.6 / 96.2} \\
        M1 & \cmark & \cmark & \cmark & SM & \cellcolor{cgrey}{29.2 / 39.5 / 49.7} & \cellcolor{cgreen}{\textbf{70.3 / 92.4 / 97.6}} \\
    \midrule        
        DIFT    & \xmark & \xmark & \xmark & GM  & \cellcolor{cgrey}{43.7 / 53.1 / 61.3}& \cellcolor{cgrey}{26.4 / 40.4 / 50.6}\\
        DIFT.S & \xmark & \xmark & \cmark & GM  & \cellcolor{cgreen}{50.4 / 58.7 / 65.7 } & \cellcolor{cgrey}{ 32.7 / 46.4 / 55.6}\\    
    \textbf{\method-Light}  & \xmark & \cmark & \cmark & GM  &  \cellcolor{cgreen}{51.4 / 60.1 / 67.1}& \cellcolor{cgrey}{{33.2 / 49.4 / 59.1}}\\  
        M1 & \cmark & \cmark & \cmark & GM  & \cellcolor{cgreen}{\textbf{54.0 / 62.7 / 69.8}} & \cellcolor{cgrey}{53.1 / 76.8 / 85.5} \\
    \midrule
    DIFT.Uni   & \xmark & \xmark & \xmark & Uni & \cellcolor{cgrey}{43.6 / 52.7 / 60.8} & \cellcolor{cgrey}{26.4 / 40.4 / 50.6} \\
    DIFT.Uni + DINO & \cmark & \xmark & \xmark & Uni & \cellcolor{cgrey}{41.9 / 51.3 / 60.0} & \cellcolor{cgrey}{58.7 / 82.9 / 90.7} \\
        M2        & \xmark & \xmark & \cmark & Uni &\cellcolor{cgreen}{50.5 / 58.9 / 65.9} & \cellcolor{cgreen}{31.8 / 45.6 / 55.4} \\ 
        M3        & \xmark & \cmark & \cmark & Uni & \cellcolor{cgreen}{50.0 / 59.0 / 66.5} & \cellcolor{cgreen}{60.8 / 82.8 / 90.4 } \\
        M4        & \cmark & \xmark & \cmark & Uni & \cellcolor{cgreen}{53.0 / 61.8 / 69.0 } & \cellcolor{cgreen}{53.9 / 78.1 / 88.2} \\
        \textbf{\method} & \cmark & \cmark & \cmark & Uni & \cellcolor{cgreen}{51.7 / 61.0 / 68.5} & \cellcolor{cgreen}{70.2 / 91.3 / 97.0 }\\    
        
\bottomrule	
\end{tabular}
}
 \vspace{-4pt}
\caption{\textbf{\method Ablation Study.} We ablate different components of proposed model on Aachen~\cite{sattler2018benchmarking}  for geometric matching and PF-Willow~\cite{ham2016pfwillow} for semantic matching using the same metrics defined in the previous sections. We denote their descriptor types using \textbf{SM/GM/Uni} that stand for semantic/geometric/unified features. We use
 \colorbox{cgreen}{green} cells for evaluations on a supervised matching task and \colorbox{cgrey}{gray} on zero-shot matching tasks. 
} 
\vspace{-0.1cm}
\label{tab:model_ablation}

\end{table}

\begin{table*}[t!]
\centering
\resizebox{16cm}{!} {
\begin{tabular}{lccccccccc}
    \toprule
              & & & \multicolumn{2}{c}{\textcolor{brown}{\textit{Geometric}}}  & \multicolumn{2}{c}{\textcolor{brown}{\textit{Semantic}}} & \multicolumn{2}{c}{\textcolor{brown}{\textit{Temporal}}}& \\
                 & Single &   Corres. & \multicolumn{2}{c}{Aachen} & \multicolumn{2}{c}{PF-Willow} & \multicolumn{2}{c}{TapVid-Davis} & Average  \\         
          Method & Desc &  Sup.& AUC$_{@5/10/20}(\uparrow)$ & Avg$(\uparrow)$ & PCK$_{@0.05/0.1/0.15}(\uparrow)$ & Avg$(\uparrow)$ & PCK$_{@0.05/0.1/0.15}(\uparrow)$ & Avg$(\uparrow)$ & Score$(\uparrow)$\\
    \midrule
        DISK~\cite{tyszkiewicz2020disk}  & \cmark & GM      & \cellcolor{cgreen}{48.9 / 57.5 / 64.6} & 57.0 &\cellcolor{cgrey}{10.2 / 17.0 / 23.1}  & 16.8 &  \cellcolor{cgrey}{57.0 / 61.7 / 65.0} & 61.2 & 45.0  \\
        
        XFeat~\cite{potje2024xfeat} &  \cmark & GM          & \cellcolor{cgreen}{36.1 / 45.9 / 55.1} & 45.7 &\cellcolor{cgrey}{25.7 / 40.0 / 48.8} & 38.2 &  \cellcolor{cgrey}{63.3 / 71.4 / 77.1} & 70.6 & 51.5 \\
        MASt3R.E~\cite{leroy2024mast3r} &  \cmark & GM      & \cellcolor{cgreen}{31.2 / 41.3 / 51.3} & 41.3 &\cellcolor{cgrey}{24.0 / 42.1 / 54.7} & 40.3 &  \cellcolor{cgrey}{75.2 / 83.8 / 87.9} & 82.3 &  54.6 \\
    \midrule        
        DIFT~\cite{tang2023dift}        & \xmark &  \xmark & \cellcolor{cgrey}{43.7 / 53.1 / 61.3}  & 52.7 & \cellcolor{cgrey}{55.7 / 85.1 / 92.9}   & 77.9  & \cellcolor{cgrey}{79.7 / 86.7 / 90.5} & 85.6 & 72.1 \\
        
        \textbf{\method-Light}     & \xmark & GM+SM &  \cellcolor{cgreen}{51.4 / 60.1 / 67.1}  &\underline{59.5} &  \cellcolor{cgreen}{69.0 / 90.6 / 96.2} & \underline{85.3}  & \cellcolor{cgrey}{78.7 / 86.3 / 90.2} & 85.1 & \underline{76.6}  \\ 
    \midrule        
        DINOv2~\cite{oquab2023dinov2}      & \cmark & \xmark & \cellcolor{cgrey}{17.2 / 26.1 / 36.4}   & 26.6  & \cellcolor{cgrey}{43.8 / 75.4 / 86.1}    & 68.4  &  \cellcolor{cgrey}{83.2 / 89.7 / 92.0} & 88.3 & 61.1 \\
        DIFT.Uni +DINOv2   & \cmark &  \xmark & \cellcolor{cgrey}{41.9 / 51.3 / 60.0}  & 51.1 &  \cellcolor{cgrey}{58.7 / 82.9 / 90.7}  & 77.4  & \cellcolor{cgrey}{86.4 / 91.6 / 93.5} & \underline{90.5} & 73.0  \\ 
        \textbf{\method} & \cmark &  GM+SM & \cellcolor{cgreen}{\textbf{51.7 / 61.0 / 68.5}} & \textbf{60.4}  & \cellcolor{cgreen}{\textbf{70.2 / 91.3 / 97.0}} & \textbf{86.2}  & \cellcolor{cgrey}{\textbf{87.8 / 93.5 / 95.5}} & \textbf{92.3} &  \textbf{79.6 }\\

    \bottomrule	
\end{tabular}
}
\caption{\textbf{Towards Matching Anything with A Unified Feature.} We compare ourselves to various feature models across geometric, semantic and temporal matching and compute the ranking of each method for each task and averaged over tasks. We show that \method is able to achieve the topk averaged ranking among all types of methods using a single feature for matching anything.}
\label{tab:towards_matching_anything}
\vspace{-0.1cm}
\end{table*}

\subsection{Ablations}

We perform ablation studies on Aachen~\cite{sattler2018benchmarking} and PF-Willow~\cite{ham2016pfwillow} for geometric and semantic matching, respectively. 
In \cref{tab:model_ablation}, we present intermediate variants that evolve from DIFT baseline towards our final \method. We focus on studying the impact of four design choices: (i) correspondence supervision, (ii) feature fusion between semantic and geometric features, (iii) leveraging DINOv2 and (iv) using separate semantic (SM) and geometric (GM) descriptors versus a unified (Uni) feature for both tasks. We assign each baseline a name for easy reference.

\PAR{Impact of correspondence supervisions.} We add the same number of self-attention layers (as in \method) to process the original DIFT semantic and geometric descriptors and supervise them accordingly using the same semantic and geometric supervisions individually. We name this variant DIFT.S.
As shown in \cref{tab:model_ablation}, the geometric supervision leads to improved performance both on geometric and semantic matching, verifying that a general improvement in matching capability was gained with geometric supervision. 
While supervised semantic DIFT descriptor also shows clear improvement on semantic matching, it leads to worse geometric matching performance, indicating the loss of generalization capability in its feature potentially due to the limited semantic matching data.

\PAR{Impact of dynamic feature fusion.}
After turning on our proposed fusion module (\cf \cref{subsec:achitecture}), \method-Light is able to further improve the accuracy on top of DIFT.S when being evaluated on both supervised and unsupervised semantic and geometric matching tasks. While semantic and geometric features contain information to support each other, it is not trivial to extract and fuse them to realize the mutual helping goal. For example, naively concatenating DIFT semantic and geometric features as in DIFT.Uni, or DIFT supervised features as in M2, both lead to a big drop in semantic matching performance compared to using those feature individually.
In contrast, we show that with the help of feature fusion, semantic and geometric features not only improve themselves as in \method-Light, but also become more cooperative and consistent with each other when being concatenated as in M3.
The above experiments fully demonstrate that our proposed feature fusion module enables effective extraction and fusion of helpful information from the semantic and geometric features into each other, leading to enhanced feature matching accuracy.

\PAR{Role of DINOv2.}
As shown in \cref{tab:model_ablation}, M1, DIFT.Uni+DINO, \method building on top of DINOv2, achieve constant improvement on both geometric and semantic matching  performance compared to their baselines  \method-Light, DIFT.Uni and M3.
Such conclusion is consistent with our discussion in \cref{sec:exp-temporal_matching}, showing that DINOv2 provides interesting complementary knowledge to DIFT as well as our supervised \method-Light, to significantly boost their general matching capabilities.

\PAR{A unified feature.} 
As shown in the upper two parts of \cref{tab:model_ablation}, using only the semantic or geometric descriptor, it is hard to achieve a good performance on both tasks. Among those, M1 geometric descriptor is the most promising feature that achieves the best geometric matching performance with proper generalization on semantic matching. 
However, unifying the semantic and geometric feature of M1 into one as in \method largely improves its performance on semantic matching, with a slight drop at geometric matching accuracy, achieving the best balance between the two matching tasks.
We further evaluate \method in the next section towards our end goal.

\subsection{Towards Matching Anything}
Keeping multiple versions of feature descriptors for an image is not effective in general. Therefore, we aim at pursuing a foundation feature model that produces a single feature descriptor that is designed for \textit{matching anything}.
In this section, we thoroughly evaluate the state-of-the-art feature models across the three matching tasks, \ie, geometric, semantic, and temporal matching.
As shown in \cref{tab:towards_matching_anything}, geometric features are not able to perform semantic matching well and have limited generalization ability on temporal matching. While the unsupervised foundation feature DIFT shows promising matching capability generalizing across three tasks, it requires different descriptors to handle different tasks and has a clear gap compared to task-specific best performing models. 
Our method, \method, building on top of the feature knowledge learned in DIFT and DINOv2, further enhanced with precise correspondence supervision and supported by a careful fusion mechanism, for the first time, outperforms all other methods across all tasks, using only a single feature.

\section{Conclusion}
\label{sec:conclusion}

In this work, we introduce a new vision challenge: achieving \textit{match-anything} capability with a single, unified feature representation. 
We propose \method, a novel feature model that harnesses existing correspondence supervision resources to narrow the accuracy gap between foundational features and task-specific supervised methods, while preserving generalization across diverse correspondence tasks. 
By incorporating limited, high-quality supervision, we take a significant step toward eliminating the need for task-specific feature descriptors, moving closer to universal matching features. This approach has direct implications for applications relying on robust correspondence, including 3D reconstruction, tracking and localization, image retrieval, and image editing.

\PAR{Limitations.} Our experiments reveal that while features derived from foundation models capture rich information, they still face challenges in resolution precision for fine-grained geometric matching and are often not optimized for runtime efficiency. We encourage future work to address these limitations for broader applicability.

\appendix
In this supplementary document, we describe model training details in \cref{sec:supp_training}, and provide more evaluation results in \cref{sec:supp_more_eval}. In addition, we present qualitative visualization for geometric, semantic and temporal matches across different methods in \cref{sec:supp_visual}.

\section{Supervision and Training Details}
\label{sec:supp_training}

\PAR{Geometric matching supervision.} 
We train geometric descriptors with ground-truth geometric correspondences as previous local features~\cite{detone2018superpoint,potje2024xfeat,revaud2019r2d2}.
We leverage the dual-softmax loss function proposed in ~\cite{potje2024xfeat} which employs the negative log-likelihood loss over matching probabilities from mutual directions.
Given an image pair $I^{a}$ and $I^{b}$ with $M$ ground-truth geometric correspondences, we first subsample sparse geometric descriptors $X_g^a$ and $X^{b}_g \in R^{M \times D_g}$ located at keypoints with ground truth annotations from the extracted dense geometric descriptors $F^{a}_{g}$ and $F^{b}_{g}$. 
We then compute the similarity matrix $S\in R^{M \times M}$ from two sets of sparse descriptors, \ie, $S=X^{a}_g(X^{b}_g)^T$, and compute the geometric loss defined as:
\begin{align}
    L_{geo} = &-\sum_i\mathtt{log}(\mathtt{softmax}_r(S)_{ii}) \nonumber \\ &- \sum_i\mathtt{log}(\mathtt{softmax}_r(S^{T})_{ii}),
    \label{eq:desc}
\end{align}
where we apply $\mathtt{softmax}$ from both matching directions over the similarity matrix.

\PAR{Semantic matching supervision.} 
Similar to geometric matching supervision, we train semantic descriptors with ground-truth semantic matches which are sparsely annotated by human. Thus, we subsample sparse semantic descriptors $X^a_s\in R^{M \times D_s}$ and $X^b_s \in R^{M \times D_s}$ at keypoint locations with ground truth.
We adopt the commonly used the CLIP contrastive loss~\cite{radford2021clip} $f_{cl}$ defined as:
\begin{align}
    f_{cl} = f_{ce}(\tau X^{a}_s(X^{b}_s)^T, \mathcal{O}) + f_{ce}(\tau X^{b}_s(X^{a}_s)^T, \mathcal{O}),
    \label{eq:sem_cl}
\end{align}
where $f_{ce}$ is the CrossEntropy loss and $\tau$ is the scale parameter. $\mathcal{O} = (0, 1, ..., M - 1)^T$ is the ground-truth labels with $M$ classes.
The sparse contrastive loss, however, only minimizes the distances between positive pairs and ignores the distances between negative pairs. To compensate for that, an additional dense semantic flow loss~\cite{lee2019sfnet} is adopted as
\begin{align}
L_{flow} = &\sum_i||(p^{a}_i - (\hat{p}^{a}_i + \epsilon))||_2 + \nonumber \\&\sum_i||(p^{b}_i - (\hat{p}^{b}_i + \epsilon))||_2 \,\,,
\label{eq:sem_flow}
\end{align}
where $\epsilon$ is the Gaussian noise with mean of 0 and standard variance of 25, $\hat{p}^{a}_i$ is the ground-truth correspondence, and $p^{a}_{i} = \sum_{q}m_i(q)q$ is the predicted correspondence.
$p^{a}_{i}$ is the average of all positions $q=(u,v)$ of $F^{b}_s$ weighted by matching probability $m_i(q)$ between descriptor $X^{a}_{s,i}$ at the index of $i$ and $F^{b}_{s,q}$. 
The matching probability $m_i(q)$ is the normalized similarity value between $X^{a}_{s,i}$ and $F^{b}_{s,q}$ and is computed as:
\begin{align}
    m_i(q)=\frac{exp(\frac{X^{a}_{s,i}(F^{b}_{s,q})^T}{\beta})}{\sum_q^{'}exp(\frac{X^{a}_{s,i}(F^{b}_{s,q^{'}})^T}{\beta})} \,\,,
    \label{eq:sem_matching}
\end{align}
where $\beta$ is the temperature.

The optical flow loss enforces semantic descriptors to maximizes the distances between negative pairs while minimizing the distances between positive pairs. The total loss for semantic matching is the combination of sparse contrastive loss $L_{cl}$ and dense flow loss $L_{flow}$:
\begin{align}
L_{sem} = w_{cl}L_{cl} + w_{flow}L_{flow},
\label{eq:sem_loss}
\end{align}
where $w_{cl}$ and $w_{flow}$ are weights balancing the two losses.

\PAR{Training data.}
We train our model using both geometric and semantic datasets, balancing samples across each dataset to ensure even representation. 
For geometric matching supervision, we use the ScanNet~\cite{dai2017scannet} and MegaDepth~\cite{li2018megadepth} datasets adopting dataset splits used in \cite{sun2021loftr} leading to approximate $15$k indoor sequences from  ScanNet and 441 outdoor sequences from MegaDepth. We use the ground-truth poses and depth maps to generate correspondences for training. 
For semantic matching  supervision, we use PF-PASCAL~\cite{ham2017pfpascal}, SPair-71k~\cite{min2019spair}, and AP-10k~\cite{yu2021ap} as in~\cite{zhang2024geosddino}. PF-PASCAL includes 2941 training pairs from 20 object categories. SPair-71k offers 53k training pairs across 18 categories with high intra-class variation. AP-10k provides 10k images across 23 categories, with an additional 261k pairs generated for semantic training.

\PAR{Training schema.} 
To properly train \method, we adopt a multi-stage training schema.
Empirically, we found geometric descriptors require more iterations to be trained properly. This is likely to be caused by the imbalanced number of  available annotated data, \ie, we have more geometric samples than semantic samples. Training too long on limited semantic matching correspondences harms generalization. 
Therefore, to compensate the data imbalance, we
1) first train the model purely on geometric matching with frozen semantic features using $L_{geo}$, and 2) next jointly train both geometric and semantic descriptors on geometric and semantic matching using a weighted combination of both supervisions as:
\begin{align}
    L_{total} = L_{geo} + w_{sem}L_{sem}.
    \label{eq:total}
\end{align}

\PAR{Implementation details.}
\method is implemented on PyTorch~\cite{paszke2019pytorch} with 8 blocks consisting of both self and cross attention layers. The hidden size of the self and cross attention layer is 512 and the number of head is 8. The dimension of final geometric and semantic descriptors is 256 and 768. The patch size $p$ used to patchify geometric and semantic features is set to $2$ for both geometric and semantic features. In the training process, hyper-parameters of $\tau$ (Eq.~\ref{eq:sem_cl}), $\beta$ (Eq.~\ref{eq:sem_matching}), $w_{cl}$ (Eq.~\ref{eq:sem_loss}), $w_{flow}$ (Eq.~\ref{eq:sem_loss}), and $w_{sem}$ (Eq.~\ref{eq:total}) are set to 0.02, 14.3, 1.0, 1.0, 0.1, respectively.

We train \method using AdamW~\cite{loshchilov2017adamw} optimizer with weight decay of $1\times 10^{-3}$ and initial learning rate of $1\times 10^{-4}$ on 4 H100 GPUs for 220k iterations in total with 150k iterations at the first stage. The learning rate is reduced to $5\times 10^{-5}$ and $2\times 10^{-5}$ after 100k and 150k iterations. The batch size is set to 24 and 48 for the first and second stage training, respectively. All images are sized to $512\times512$ in the training process.

\begin{table}[t!]
\centering
\resizebox{8cm}{!} {
\begin{tabular}{lcc}
    \toprule
    && TAPVid-Davis~\cite{doersch2022tapvid} \\
         Method & Supervision &  PCK$_{@0.01/0.05/0.1}$ 
 \\
 \toprule
    DIFT (geo)~\cite{tang2023dift} &  \xmark &  75.6 / 82.6 / 86.9 \\
    DIFT (sem)~\cite{tang2023dift} &  \xmark &  71.9 / 81.4 / 86.4\\
    \method-Light (geo) & GM+SM & 75.7 / 82.8 / 87.0 \\
    \method-Light (sem) & GM+SM & 64.9 / 77.9 / 84.3 \\
    \midrule
    
    DINOv2~\cite{oquab2023dinov2} & \xmark & 83.2 / 89.7 / 92.0 \\
    DIFT.Uni~\cite{tang2023dift} & \xmark & 79.7 / 86.7 / 90.5 \\
    DIFT.Uni + DINOv2~\cite{tang2023dift,oquab2023dinov2} & \xmark & \underline{86.4} / \underline{91.6} / \underline{93.5} \\
    \method-Light.Uni & GM+SM &  78.7 / 86.3 / 90.2 \\  
    \method & GM+SM &  \textbf{87.8} / \textbf{93.5} / \textbf{95.5} \\ 
	\bottomrule	
\end{tabular}
}
\caption{\textbf{Ablation Study on Temporal Matching.} We report the Percentage of Correct Keypoints (PCK) under different thresholds. The \textbf{best} and \underline{second-best} results are highlighted.}

\label{tab:ab_temp_matching}

\end{table}

\begin{table*}[t!]
\centering
\resizebox{16cm}{!} {
\begin{tabular}{lccccccccc}
    \toprule
              & & & \multicolumn{2}{c}{\textcolor{brown}{\textit{Geometric}}}  & \multicolumn{2}{c}{\textcolor{brown}{\textit{Semantic}}} & \multicolumn{2}{c}{\textcolor{brown}{\textit{Temporal}}}& \\
                 & Single &   Corres. & \multicolumn{2}{c}{Aachen} & \multicolumn{2}{c}{PF-Willow} & \multicolumn{2}{c}{TapVid-Davis} & Average  \\         
          Method & Desc &  Sup.& AUC$_{@5/10/20}(\uparrow)$ & Avg$(\uparrow)$ & PCK$_{@0.05/0.1/0.15}(\uparrow)$ & Avg$(\uparrow)$ & PCK$_{@0.05/0.1/0.15}(\uparrow)$ & Avg$(\uparrow)$ & Score$(\uparrow)$\\
          
    \midrule

        \textbf{\method-Light}     & \xmark & GM+SM &  \cellcolor{cgreen}{51.4 / 60.1 / 67.1}  &\underline{59.5} &  \cellcolor{cgreen}{69.0 / 90.6 / 96.2} & \underline{85.3}  & \cellcolor{cgrey}{78.7 / 86.3 / 90.2} & 85.1 & \underline{76.6}  \\ 
         \method-Light.Uni  & \cmark & GM+SM & \cellcolor{cgreen}{50.0 / 59.0 / 66.5}   & 58.5 & \cellcolor{cgreen}{60.8 / 82.8/ 90.4}   & 78.0 & \cellcolor{cgrey}{78.7 / 86.3 / 90.2} & 85.1 &  73.9\\  
         \method-Light.Uni.S & \cmark & GM+SM & \cellcolor{cgreen}{49.9 / 58.4 / 65.4}& 57.9 & \cellcolor{cgreen}{36.8 / 53.0 / 62.4} & 50.7 & \cellcolor{cgrey}{79.1 / 85.9 / 89.5} & 84.8 & 64.5\\
        \textbf{\method} & \cmark &  GM+SM & \cellcolor{cgreen}{\textbf{51.7 / 61.0 / 68.5}} & \textbf{60.4}  & \cellcolor{cgreen}{\textbf{70.2 / 91.3 / 97.0}} & \textbf{86.2}  & \cellcolor{cgrey}{\textbf{87.8 / 93.5 / 95.5}} & \textbf{92.3} &  \textbf{79.6 }\\

    \bottomrule	
\end{tabular}
}
\caption{\textbf{Ablation study on obtaining a unified feature.} We compare different ways of obtaining a unified feature. We show that simple concatenation leads to better way to keep the learned geometric and semantic representation while adding additional joint training on the concatenated feature pushes the feature to focus more on geometric matching,  leading to significantly degraded semantic matching.}
\label{tab:supp_ablat_unify_feat}
\vspace{-0.1cm}
\end{table*}

\section{Additional Evaluations}
\label{sec:supp_more_eval}
\PAR{Temporal matching.}
We provide additional ablation study to understand the performance of different types of features on temporal matching. 
Specifically, we consider the geometric (geo) and semantic (sem) of descriptors of \method-Light and DIFT~\cite{tang2023dift} models and their unified feature version, \ie, DIFT.Uni, \method-Light.Uni.
We also consider the feature models that combine DINOv2~\cite{oquab2023dinov2}, \ie, DIFT.Uni + DINOv2 and \method.

As shown in \cref{tab:ab_temp_matching},  low-level geometric features are more important to temporal matching than high-level semantic features, \ie,  DIFT (geo) \textit{vs} DIFT (sem), and \method-Light (geo) \textit{vs} \method-Light (sem).
We also notice that geometric supervision leads to improved temporal matching, \ie, \method-Light (geo) \textit{vs} DIFT (geo).
In contrast, adding semantic supervision produces degraded temporal matching accuracy,  \ie, \method-Light (sem) \textit{vs} DIFT (sem), which shows that sparse semantic correspondence supervision across instances leads to decreased capability in establishing matches between the same instance.

The combination of geometric and semantic features contains the properties of the both features, giving better temporal matching accuracy \ie, DIFT.Uni \textit{vs} DIFT (geo), and \method-Light.Uni \textit{vs} \method-Light (geo). 
As discussed in the main paper, DINOv2 benefiting from its large-scale learning on single object-centric data, is able to well handle large viewpoint and scale changes, especially for single-object dominant scenes, leading to surprisingly superior temporal matching performance.
By combining with DINOv2 features, both DIFT.Uni and \method have significant improvement in temporal matching, \ie, \method \textit{vs} \method-Light.Uni, DIFT.Uni + DINOv2 \textit{vs} DIFT.Uni, validating object-level semantic representation learned by DINOv2 is complementary to semantic features extracted from stable diffusion models.

\PAR{Ablation on obtaining a unified feature.} In the main paper, we adopt a simple concatenation-based merging mechanism to obtain a unified feature. To further validate this design choice, we provide additional ablation study focusing on comparing different ways of unifying knowledge in feature representations. 
Specifically, we consider \textbf{\method-Light} that learns to fuse geometric and semantic features yet keeping separate descriptors for geometric and semantic matching following DIFT, \textbf{\method-Light.Uni} that combines the \method-Light geometric and semantic descriptors with concatenation-based merging, \textbf{\method-Light.Uni.S} that further supervises \method-Light.Uni with joint geometric and semantic training, as well as \textbf{\method}, our final model, that combines DINOv2 with \method-Light.Uni.

In \cref{tab:supp_ablat_unify_feat}, we show that simple concatenation-based merging (\method-Light.Uni) can effectively unify both semantic and geometric matching capabilities learned by \method-Light, giving a single feature at slight decrease in matching performance across tasks.
When we further finetune such unified feature with joint geometric and semantic matching supervsion, we observe significant drop in semantic matching performance. We consider such behavior is mainly caused by the imbalanced training data between geometric and semantic matching.
Compared to training individual descriptors, such data limitation imposes more challenges for balancing the two tasks when training a single unified descriptor.
Therefore, we finally opt for simple concatenation as our mechanism to unify different types of foundation feature representations.
It turns out to be highly effective also when combing the complementary semantic knowledge learn by DINOv2 into \method.

\section{Visualization}
\label{sec:supp_visual}
Finally, we provide visualization for different feature models through their feature similarity distribution as well as the established correspondences across different scenes. 
We compare \method-Light and \method to MASt3R.E~\cite{leroy2024mast3r}, DISK~\cite{tyszkiewicz2020disk}, DINOv2~\cite{oquab2023dinov2}, DIFT~\cite{tang2023dift} and the supervised DIFT (DIFT.S).

\begin{figure*}[t!]
    \centering
    \includegraphics[width=\linewidth]{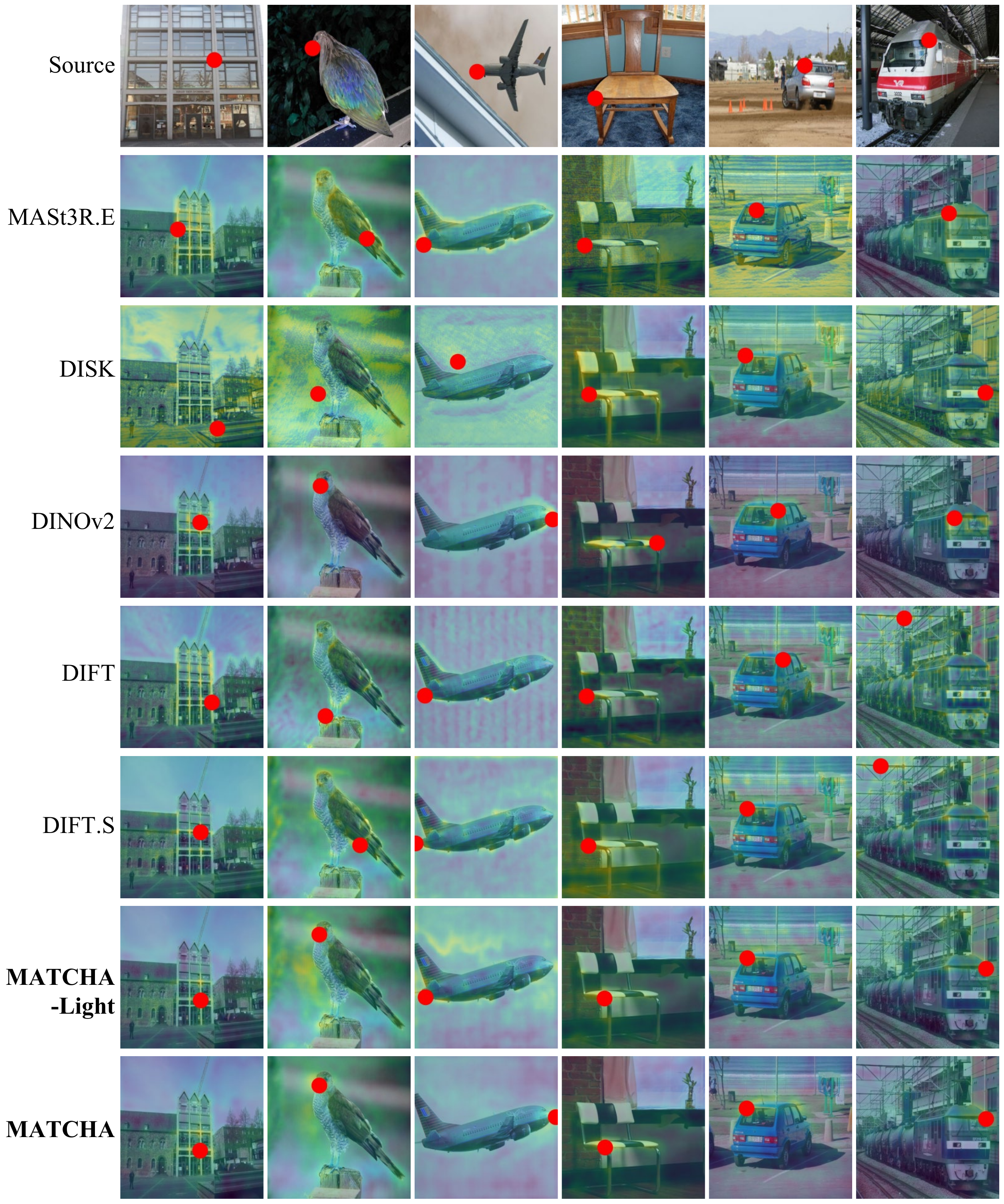}
    \caption{\textbf{Visualization of heatmap.} Given a source point (top), we visualize the heatmap and predicted matches of MASt3R.E~\cite{leroy2024mast3r}, DISK~\cite{tyszkiewicz2020disk}, DINOv2~\cite{oquab2023dinov2}, DIFT~\cite{tang2023dift}, DIFT.S (fully supervised version of DIFT), and our models \method-Light and \method.} 
    \label{fig:vis_heatmap_full}
\end{figure*}

\PAR{Heatmap.} 
In \cref{fig:vis_heatmap_full}, we visualize the heatmaps and predicted matches produced by different methods, starting from a given source point. The heatmaps represent the normalized cosine similarity between the features extracted at the source point and every pixel in the target image. DISK~\cite{tyszkiewicz2020disk}, as a local feature method, focuses primarily on local texture regions, often resulting in poor matches in scenes with repetitive structures or semantically similar content. MASt3R~\cite{leroy2024mast3r}, despite being trained on a larger dataset, still exhibits similar limitations, providing suboptimal matches in these challenging scenarios. DINOv2~\cite{oquab2023dinov2}, on the other hand, excels in cases with single objects, producing sharp and localized heatmaps. However, its performance degrades in the presence of noisy backgrounds or repetitive structures, where it fails to generate accurate matches.

For DIFT.Uni, DIFT.S.Uni, and \method-Light, we compute their heatmaps using concatenated geometric and semantic features. DIFT captures more low-level texture details, leading to high similarity scores in regions with repetitive patterns. DIFT.S.Uni improves upon DIFT.Uni by incorporating supervision, but it remains less robust to variations in semantic content due to its task-specific training. \method-Light, with joint training and dynamic feature fusion, addresses these issues to some extent, providing more accurate matches for both repetitive textures and semantically rich content. However, as it shares the same diffusion-based features as DIFT.Uni and DIFT.S.Uni, it struggles with ambiguity in visually similar parts of the same object, such as the head and tail of an airplane.

Finally, \method resolves these challenges by incorporating complementary object-level features from DINOv2. This integration significantly enhances its ability to disambiguate similar object parts and produce accurate matches even in complex scenes, making it the most robust method among the evaluated approaches.

\begin{figure*}[t!]
    \centering
    \includegraphics[width=\linewidth]{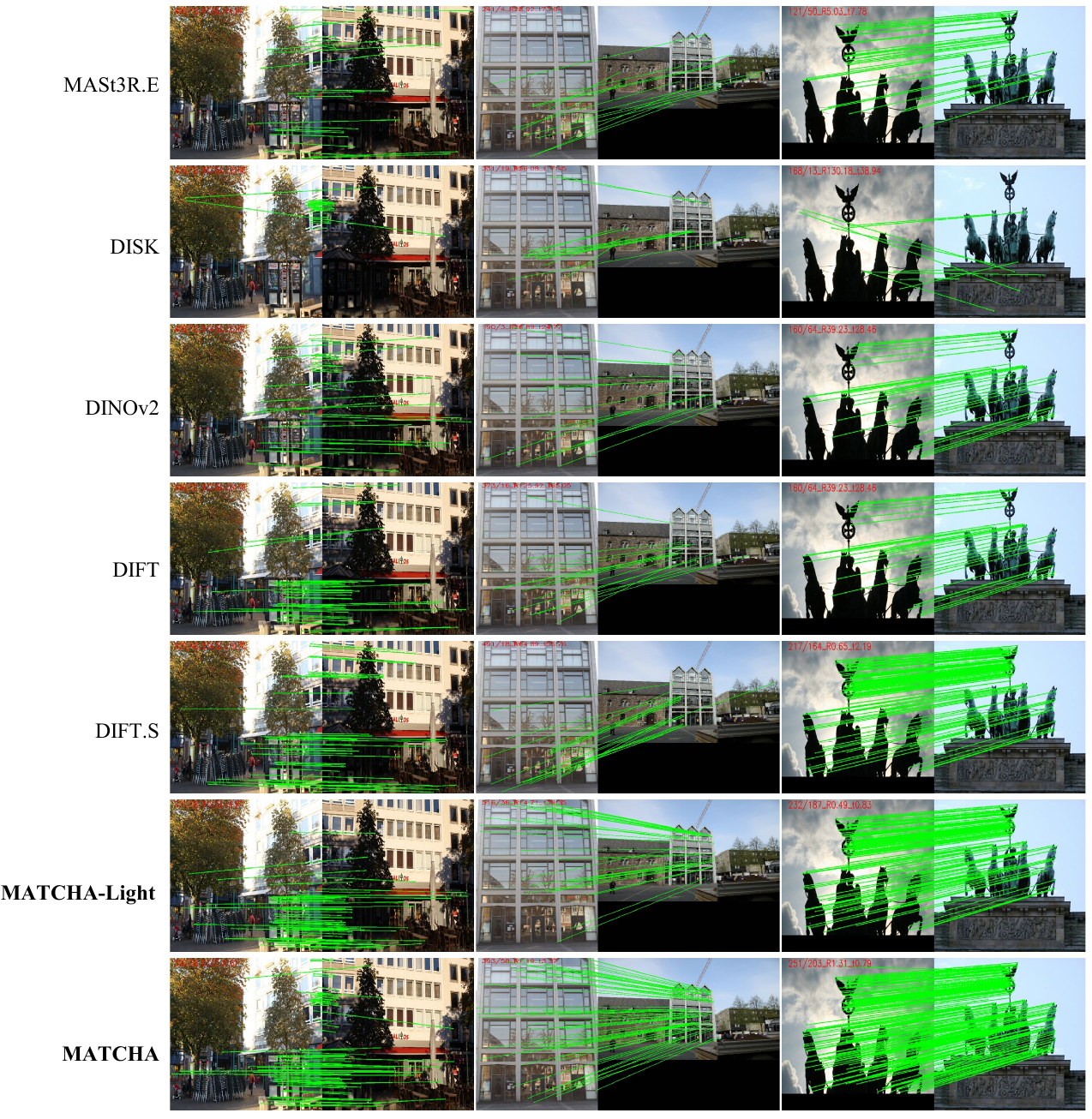}
    \caption{\textbf{Geometric matches on outdoor scenes.} We visualize the inliers after RANSAC of MASt3R.E~\cite{leroy2024mast3r}, DISK~\cite{tyszkiewicz2020disk}, DINOv2~\cite{oquab2023dinov2}, DIFT~\cite{tang2023dift}, DIFT.S (fully supervised version of DIFT), and our models \method-Light and \method. DISK produces many inliers on local patches but is not robust to repetitive structures. MASt3R and DINOv2 focus more on structures and give close performance. DIFT works better than DINOv2 especially on regions with rich textures. With geometric supervision, DIFT.S improves the performance of DIFT. \method-Light is able to find correct matches from both local patches and structures because of dynamic fusion and this ability is further enhanced by fusing features of DINOv2.} 
    \label{fig:vis_geo_matches_outdoor}
\end{figure*}

\begin{figure*}[t!]
    \centering
    \includegraphics[width=\linewidth]{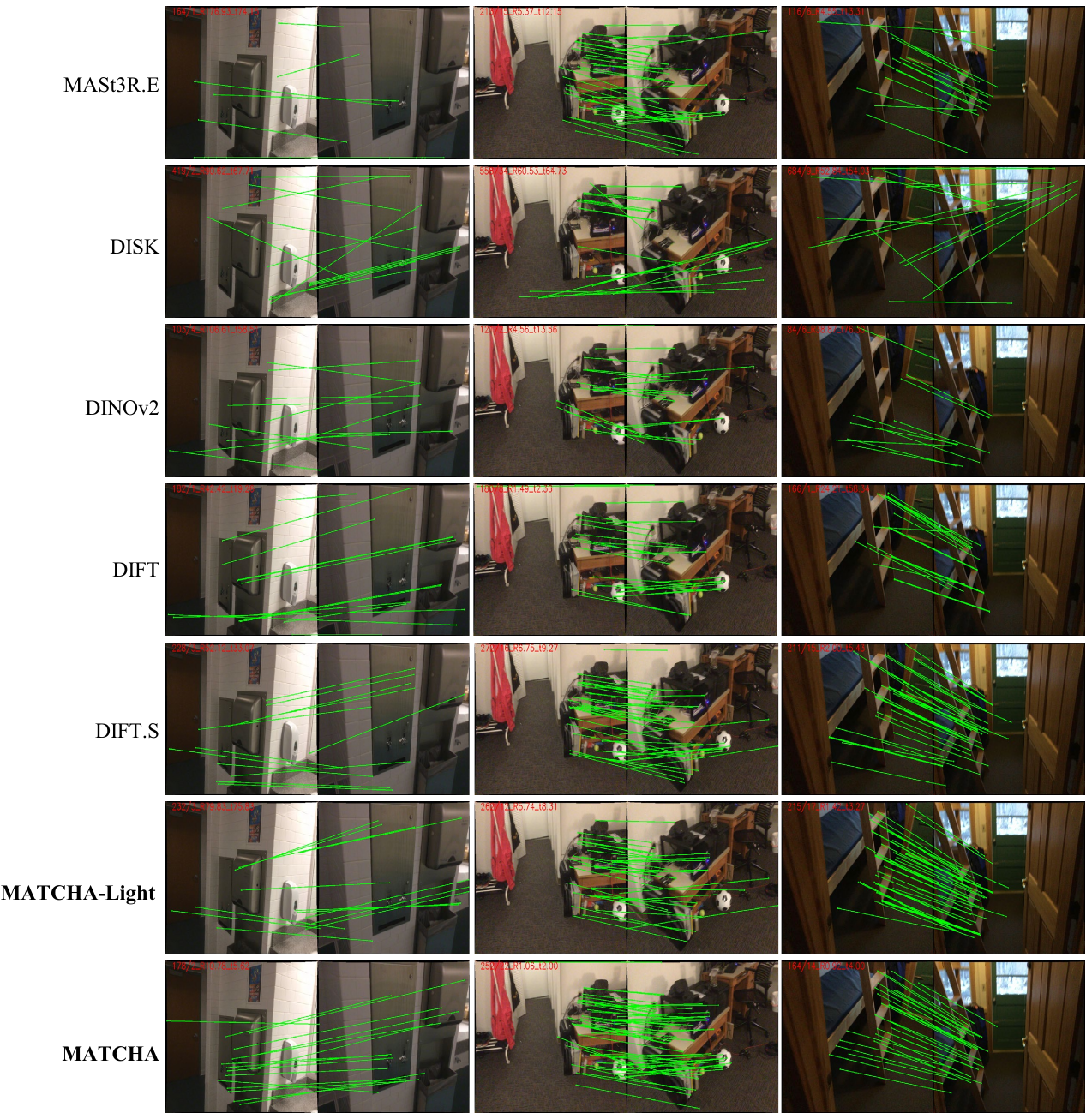}
    \caption{\textbf{Geometric matches on indoor scenes.} We visualize the inliers after RANSAC of MASt3R.E~\cite{leroy2024mast3r}, DISK~\cite{tyszkiewicz2020disk}, DINOv2~\cite{oquab2023dinov2}, DIFT~\cite{tang2023dift}, DIFT.S (fully supervised version of DIFT), and our models \method-Light and \method. Almost all previous methods fail to find sufficient inliers on scenes with repetitive structures except \method which fuses both low and high-level information. Additionally, \method is able to produce more inliers in scenes with rich textures (\textit{middle column}).} 
    \label{fig:vis_geo_matches_indoor}
\end{figure*}

\PAR{Geometric matching.} 
In \cref{fig:vis_geo_matches_outdoor} and \cref{fig:vis_geo_matches_indoor}, we evaluate geometric matching using relative camera pose estimation and RANSAC to identify inlier matches for both indoor and outdoor scenes. Geometric methods such as DISK~\cite{tyszkiewicz2020disk} and MASt3R.E~\cite{leroy2024mast3r} primarily rely on low-level texture patterns, which limits their ability to handle repetitive textures and capture high-level structures effectively. In contrast, DINOv2~\cite{oquab2023dinov2} focuses on object-level features, capturing higher-level structures but yielding sparse matches due to its limited reliance on detailed textures. DIFT strikes a balance between low- and high-level information, yet its lack of geometric supervision reduces the number of inliers compared to its supervised counterpart, DIFT.S, \method-Light, with its dynamic fusion mechanism, propagates high-level semantic knowledge to geometric features, resulting in improved inliers at both object- and patch-level. This ability is further enhanced in \method, where additional features from DINOv2 are fused, leading to the highest number of inliers among the evaluated methods.

\begin{figure*}[t!]
    \centering
    \includegraphics[width=\linewidth]{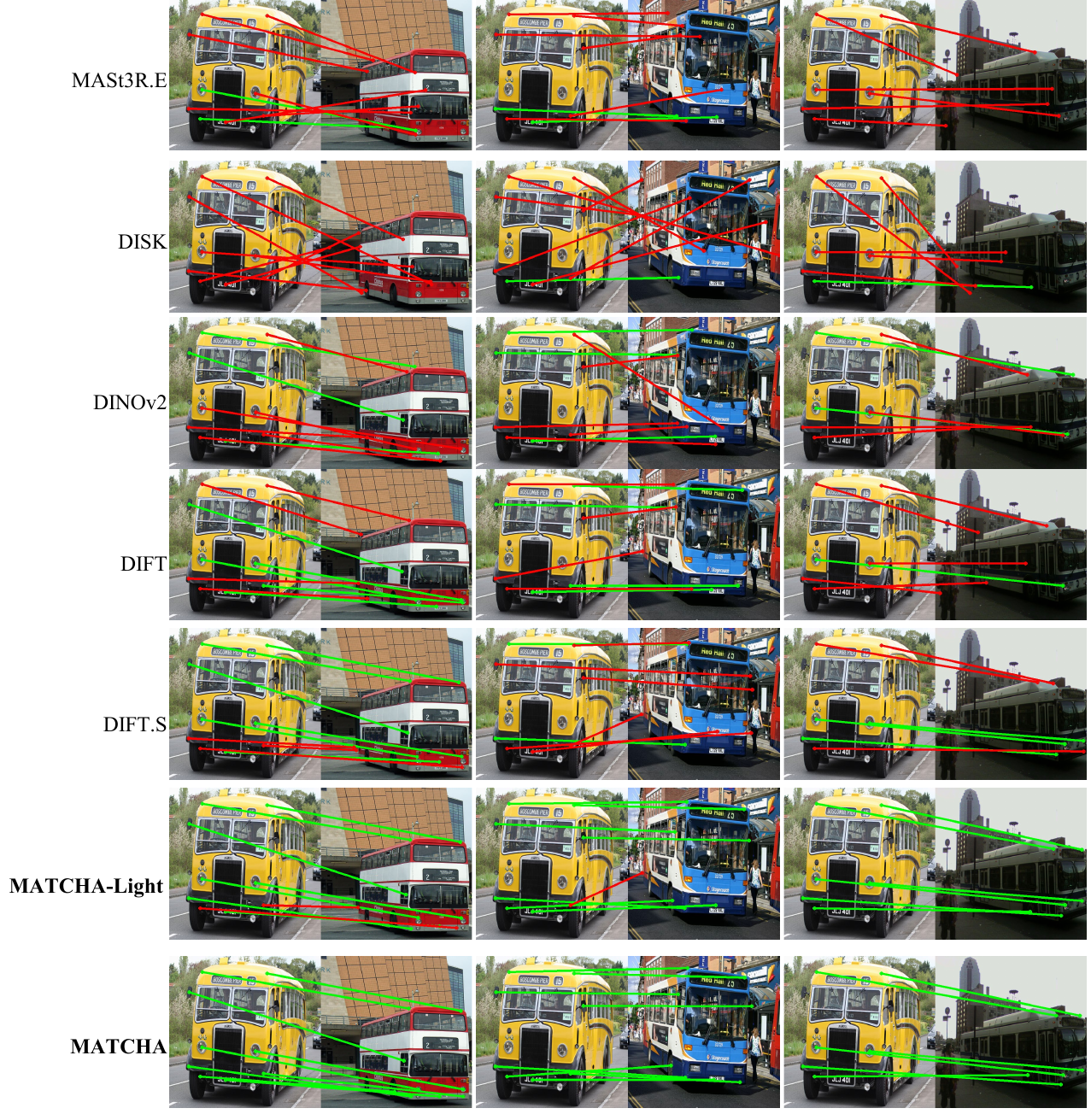}
    \caption{\textbf{Semantic matches on bus category.} We visualize the \textcolor{green}{inliers} and \textcolor{red}{outliers} of MASt3R.E~\cite{leroy2024mast3r}, DISK~\cite{tyszkiewicz2020disk}, DINOv2~\cite{oquab2023dinov2}, DIFT~\cite{tang2023dift}, DIFT.S (fully supervised version of DIFT), and our models \method-Light and \method.} 
    \label{fig:vis_sem_matches_bus}
\end{figure*}

\begin{figure*}[t!]
    \centering
    \includegraphics[width=\linewidth]{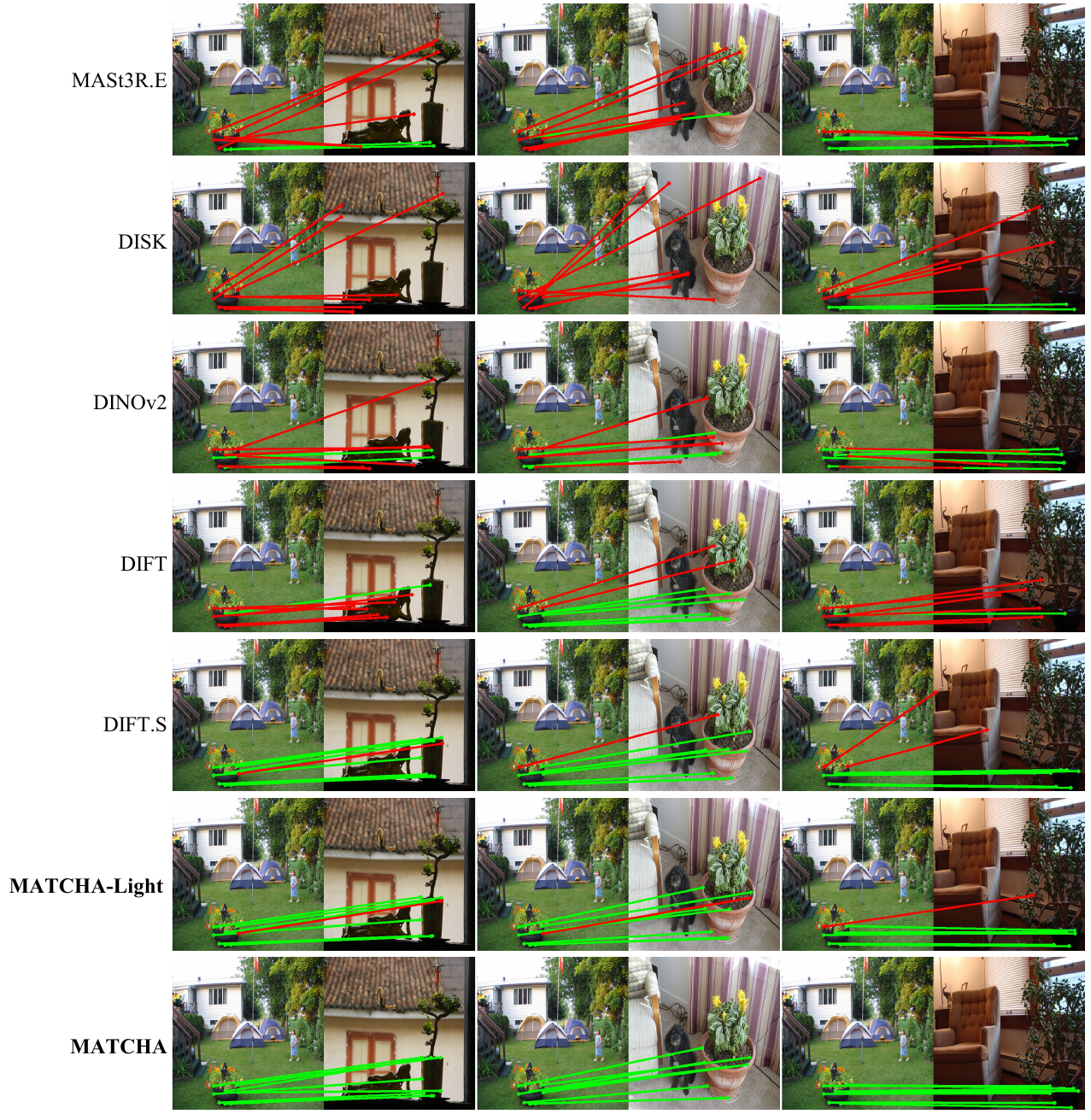}
    \caption{\textbf{Semantic matches on plant category.} We visualize the \textcolor{green}{inliers} and \textcolor{red}{outliers} of MASt3R.E~\cite{leroy2024mast3r}, DISK~\cite{tyszkiewicz2020disk}, DINOv2~\cite{oquab2023dinov2}, DIFT~\cite{tang2023dift}, DIFT.S (fully supervised version of DIFT), and our models \method-Light and \method.} 
    \label{fig:vis_sem_matches_plant}
\end{figure*}

\begin{figure*}[t!]
    \centering
    \includegraphics[width=\linewidth]{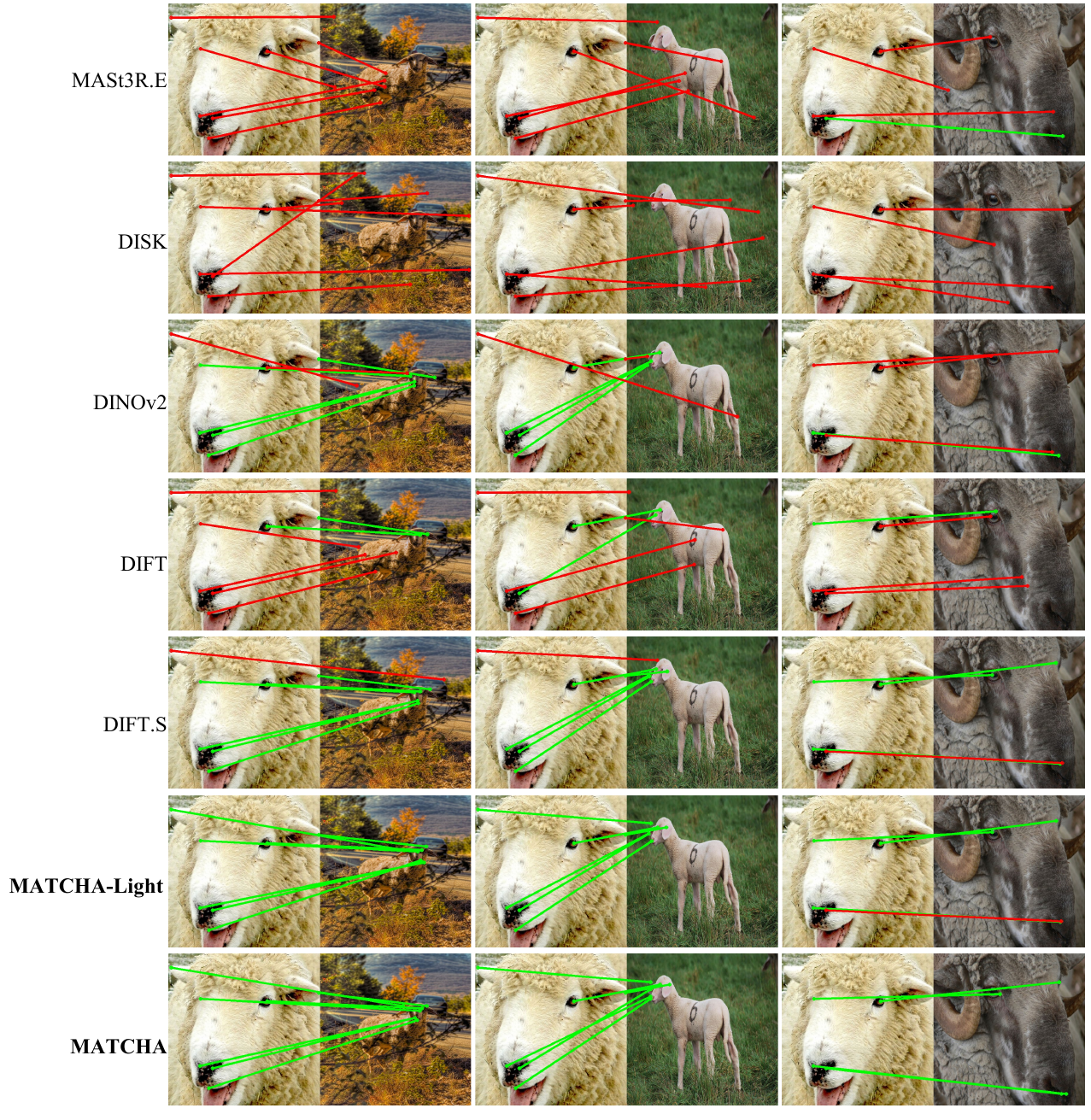}
    \caption{\textbf{Semantic matches on sheep category.} We visualize the \textcolor{green}{inliers} and \textcolor{red}{outliers} of MASt3R.E~\cite{leroy2024mast3r}, DISK~\cite{tyszkiewicz2020disk}, DINOv2~\cite{oquab2023dinov2}, DIFT~\cite{tang2023dift}, DIFT.S (fully supervised version of DIFT), and our models \method-Light and \method.} 
    \label{fig:vis_sem_matches_sheep}
\end{figure*}

\begin{figure*}[t!]
    \centering
    \includegraphics[width=\linewidth]{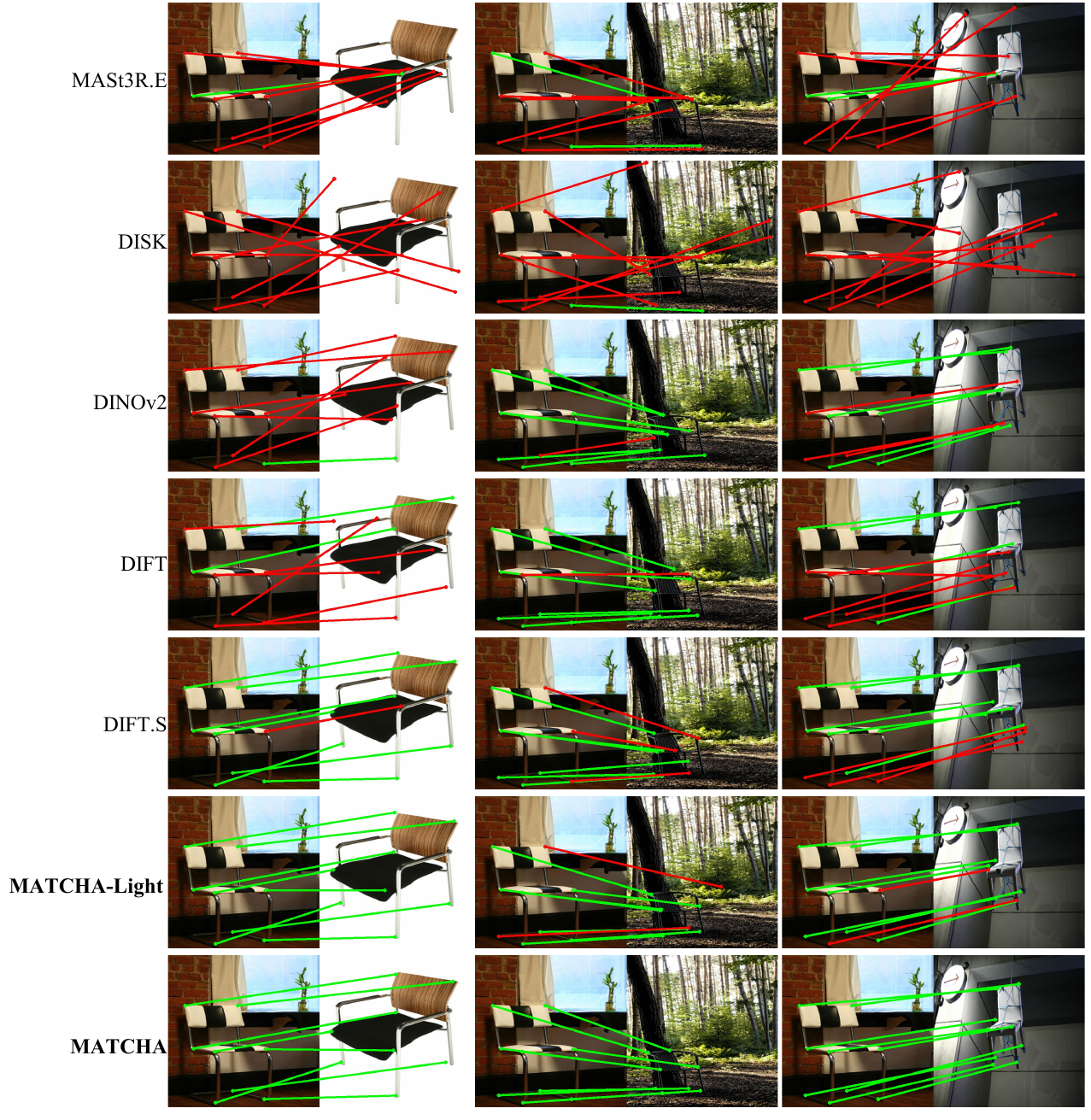}
    \caption{\textbf{Semantic matches on chair category.} We visualize the \textcolor{green}{inliers} and \textcolor{red}{outliers} of MASt3R.E~\cite{leroy2024mast3r}, DISK~\cite{tyszkiewicz2020disk}, DINOv2~\cite{oquab2023dinov2}, DIFT~\cite{tang2023dift}, DIFT.S (fully supervised version of DIFT), and our models \method-Light and \method.} 
    \label{fig:vis_sem_matches_chair}
\end{figure*}

\begin{figure*}[t!]
    \centering
    \includegraphics[width=\linewidth]{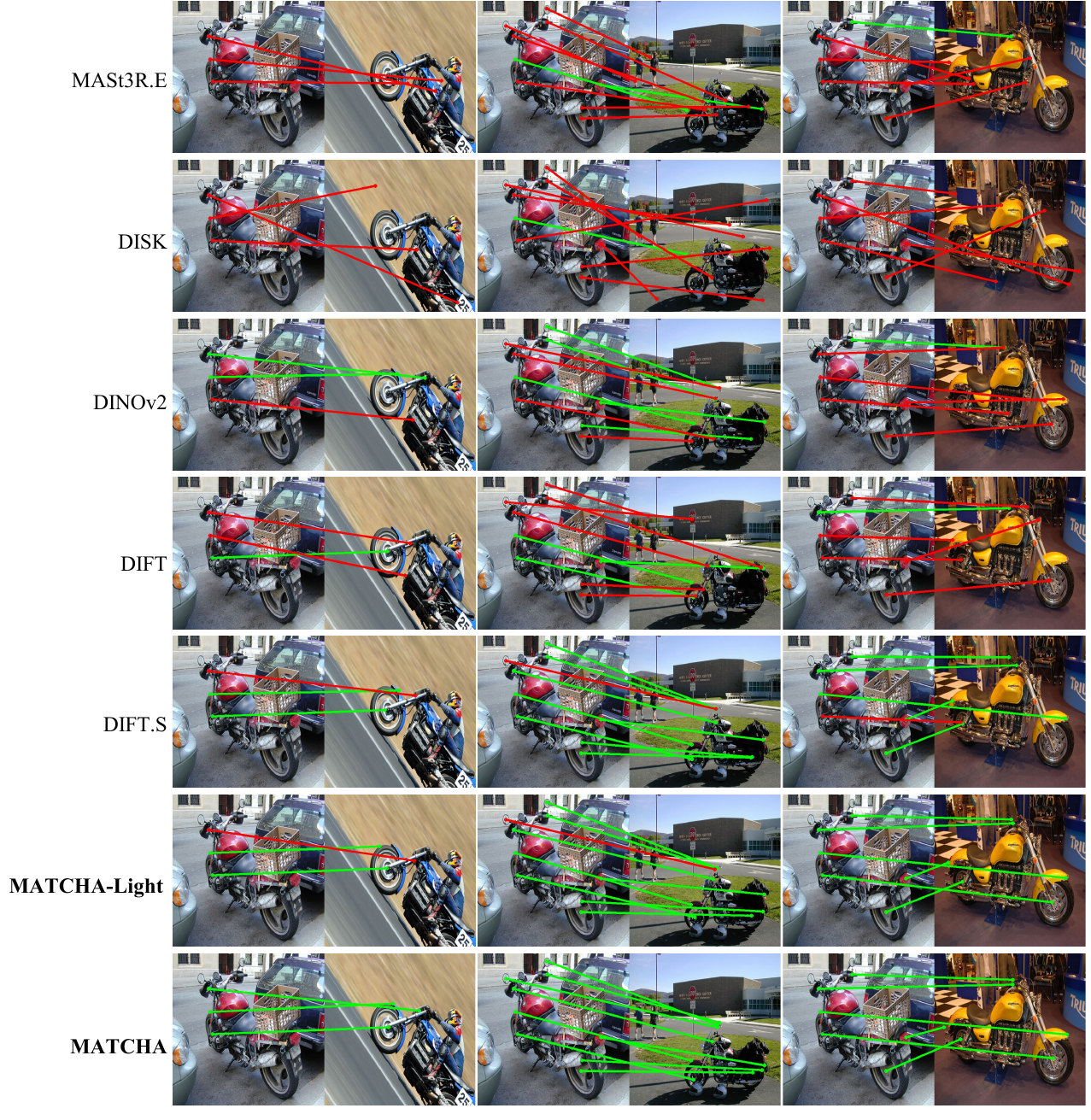}
    \caption{\textbf{Semantic matches on motorbike category.} We visualize the \textcolor{green}{inliers} and \textcolor{red}{outliers} of MASt3R.E~\cite{leroy2024mast3r}, DISK~\cite{tyszkiewicz2020disk}, DINOv2~\cite{oquab2023dinov2}, DIFT~\cite{tang2023dift}, DIFT.S (fully supervised version of DIFT), and our models \method-Light and \method.} 
    \label{fig:vis_sem_matches_motorbike}
\end{figure*}

\PAR{Semantic matching.} 
As illustrated in \cref{fig:vis_sem_matches_bus}, \cref{fig:vis_sem_matches_plant}, \cref{fig:vis_sem_matches_sheep}, \cref{fig:vis_sem_matches_chair}, and \cref{fig:vis_sem_matches_motorbike}, we visualize both \textcolor{green}{inliers} and \textcolor{red}{outliers} across various objects to assess semantic matching performance. Local geometric features, such as DISK~\cite{tyszkiewicz2020disk} and MASt3R.E, fail to establish meaningful semantic correspondences, as they primarily rely on low-level textures and patterns. 

While feature models like DINOv2, DIFT, and DIFT.S demonstrate a coarse ability to capture semantic correspondences, they often struggle with utilizing low-level details for precise local discrimination, leading to inaccuracies in challenging scenarios. In contrast, \method effectively integrates both geometric and semantic cues, achieving robust and accurate semantic matches even under extreme scale and viewpoint variations, outperforming other methods in these complex scenarios.

\begin{figure*}[t!]
    \centering
    \includegraphics[width=\linewidth]{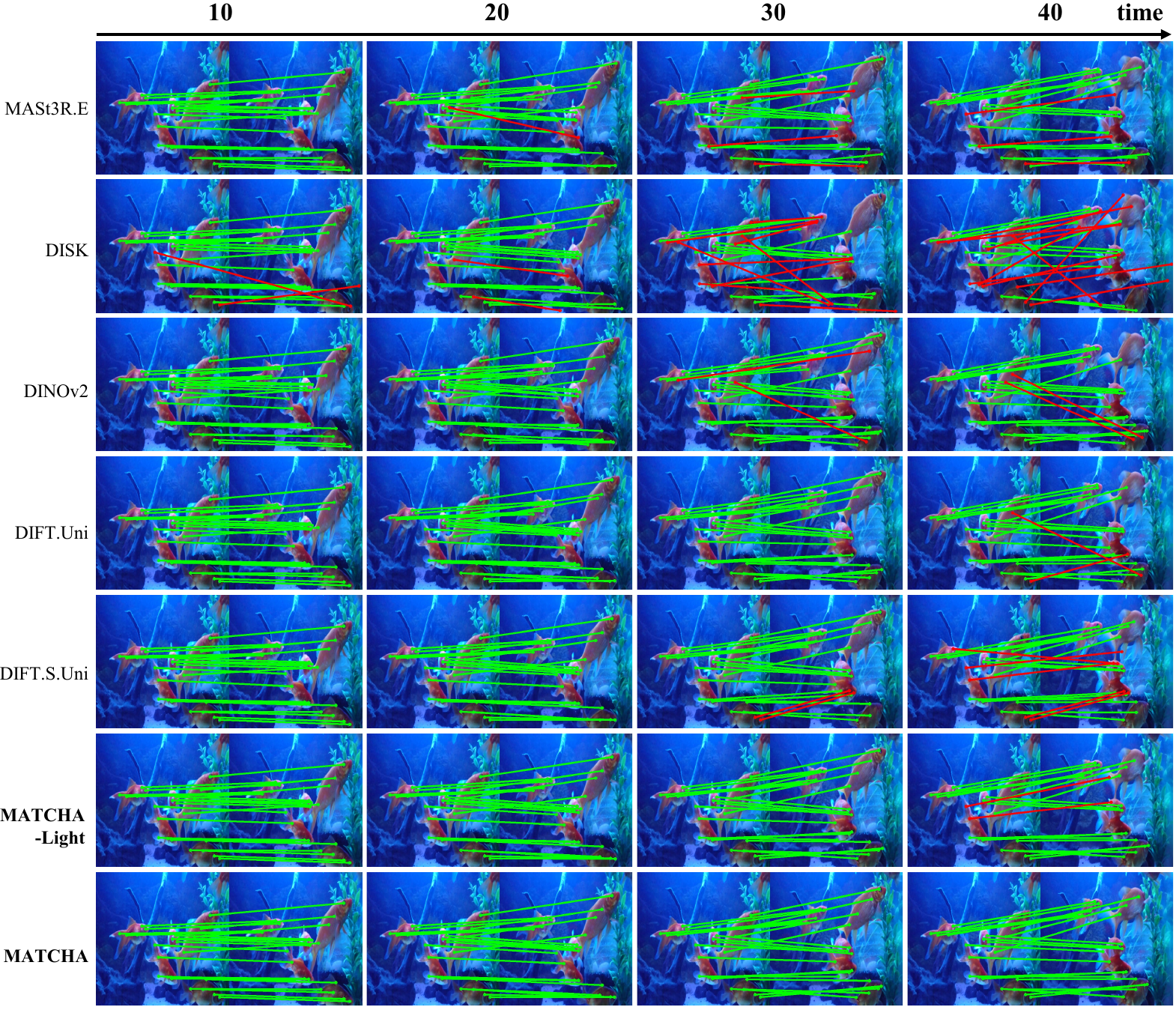}
    \caption{\textbf{Temporal matches on goldfish sequence.} We visualize the \textcolor{green}{inliers} and \textcolor{red}{outliers} of MASt3R.E~\cite{leroy2024mast3r}, DISK~\cite{tyszkiewicz2020disk}, DINOv2~\cite{oquab2023dinov2}, DIFT.Uni~\cite{tang2023dift}, DIFT.S.Uni (fully supervised version of DIFT), and our models \method-Light and \method.} 
    \label{fig:vis_tem_matches_goldfish}
\end{figure*}

\begin{figure*}[t!]
    \centering
    \includegraphics[width=\linewidth]{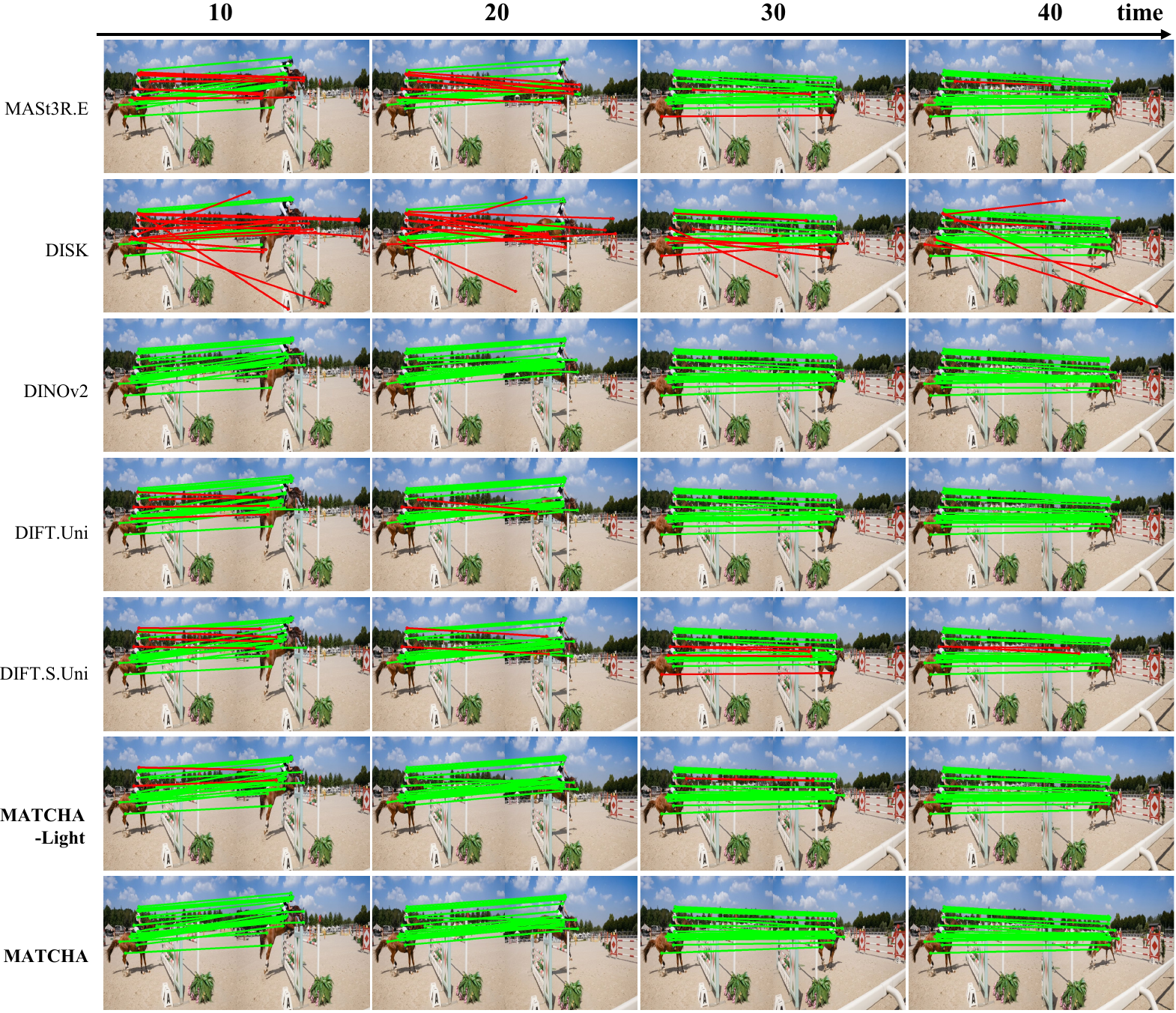}
    \caption{\textbf{Temporal matches on horsejumphigh sequence.} We visualize the \textcolor{green}{inliers} and \textcolor{red}{outliers} of MASt3R.E~\cite{leroy2024mast3r}, DISK~\cite{tyszkiewicz2020disk}, DINOv2~\cite{oquab2023dinov2}, DIFT.Uni~\cite{tang2023dift}, DIFT.S.Uni (fully supervised version of DIFT), and our models \method-Light and \method.} 
    \label{fig:vis_tem_matches_horsejumphigh}
\end{figure*}

\begin{figure*}[t!]
    \centering
    \includegraphics[width=\linewidth]{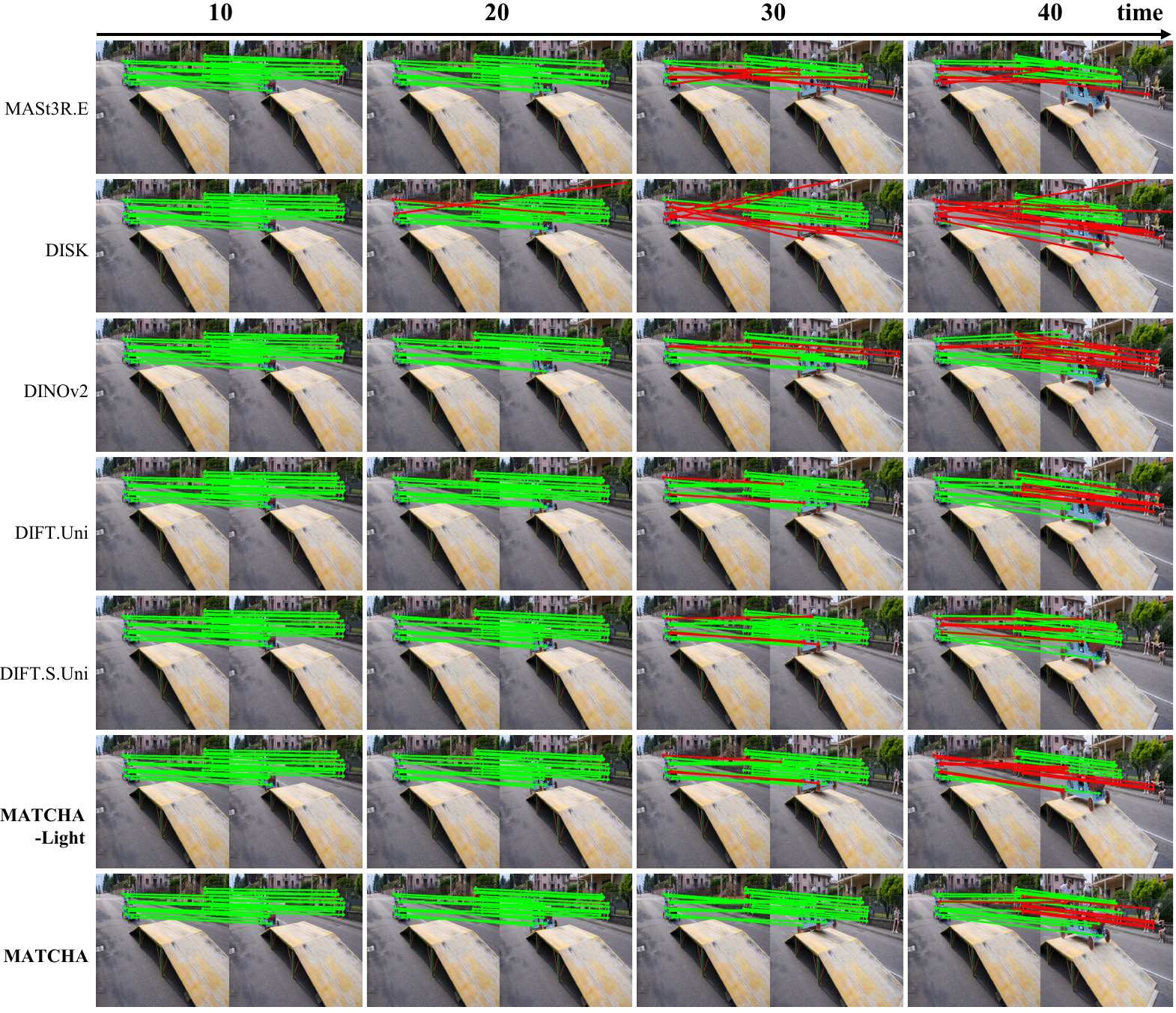}
    \caption{\textbf{Temporal matches on soapbox sequence.} We visualize the \textcolor{green}{inliers} and \textcolor{red}{outliers} of MASt3R.E~\cite{leroy2024mast3r}, DISK~\cite{tyszkiewicz2020disk}, DINOv2~\cite{oquab2023dinov2}, DIFT.Uni~\cite{tang2023dift}, DIFT.S.Uni (fully supervised version of DIFT), and our models \method-Light and \method.} 
    \label{fig:vis_tem_matches_soapbox}
\end{figure*}

\begin{figure*}[t!]
    \centering
    \includegraphics[width=\linewidth]{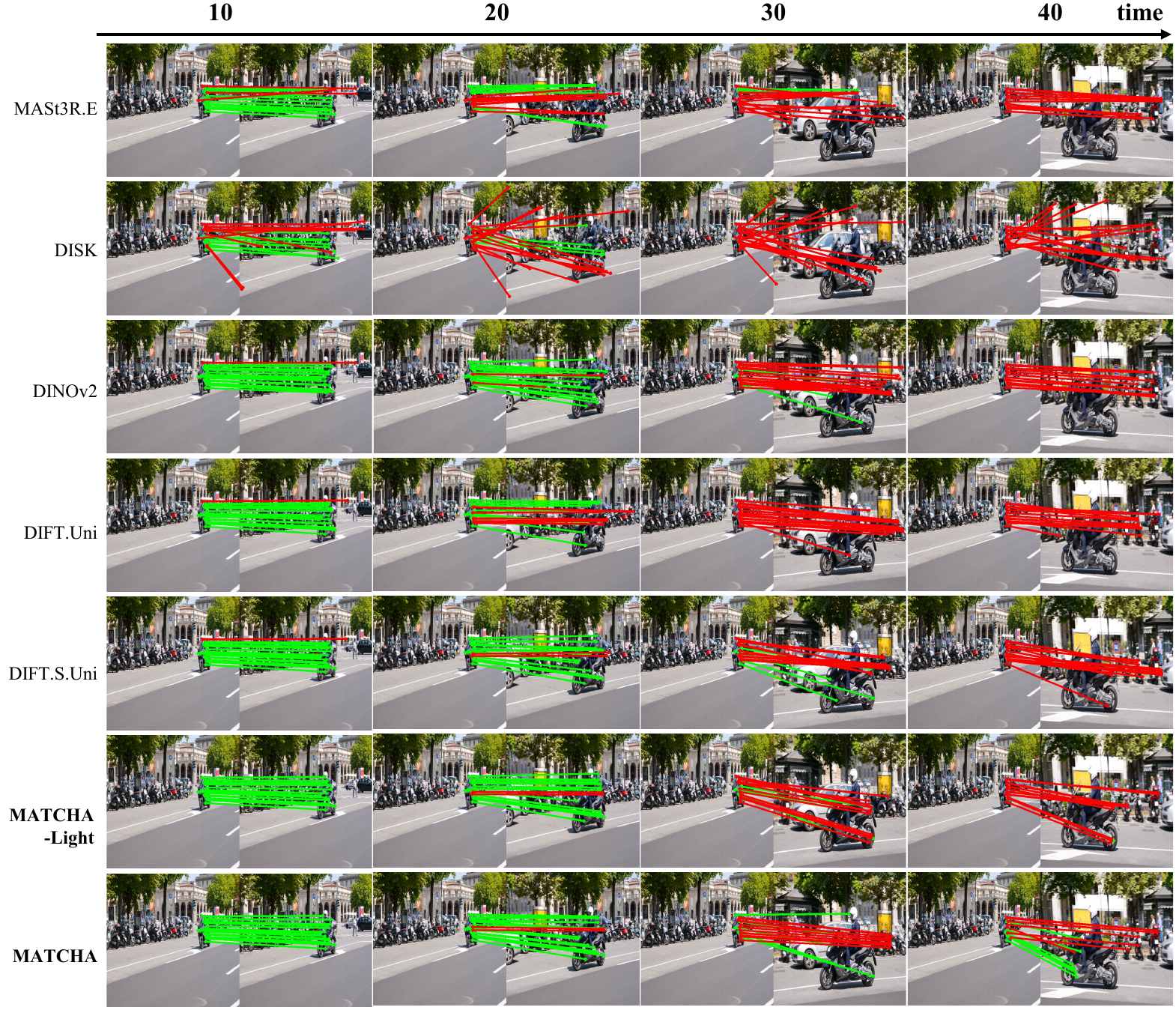}
    \caption{\textbf{Temporal matches on scooterblack sequence.} We visualize the \textcolor{green}{inliers} and \textcolor{red}{outliers} of MASt3R.E~\cite{leroy2024mast3r}, DISK~\cite{tyszkiewicz2020disk}, DINOv2~\cite{oquab2023dinov2}, DIFT.Uni~\cite{tang2023dift}, DIFT.S.Uni (fully supervised version of DIFT), and our models \method-Light and \method.} 
    \label{fig:vis_tem_matches_scooterblack}
\end{figure*}

\PAR{Temporal matching.} 
We present visualizations of temporal matches in \cref{fig:vis_tem_matches_goldfish}, \cref{fig:vis_tem_matches_horsejumphigh}, \cref{fig:vis_tem_matches_soapbox}, and \cref{fig:vis_tem_matches_scooterblack}, evaluating the performance of various methods. In addition to previously discussed baselines, we include the unified DIFT feature variants, DIFT.Uni and DIFT.S.Uni, which demonstrate improved temporal matching compared to their specific geometric or semantic descriptors.

Overall, we observe that the local feature DISK performs the worst in handling highly dynamic objects, such as a jumping horse or moving bikes, due to its reliance on low-level patterns. MASt3R.E shows marginal improvement but is still outperformed by other methods. Among all approaches, \method stands out as the most accurate and robust for temporal matching, effectively handling the challenges of dynamic scenes.

\PAR{Failure cases.} Despite its strengths, temporal matching remains a challenging task, as shown in \cref{fig:vis_tem_matches_soapbox} and \cref{fig:vis_tem_matches_scooterblack}. All methods struggle in scenarios where repetitive patterns in the background coincide with extreme scale and viewpoint changes caused by object motion. These limitations highlight the need for further research to improve the robustness and accuracy of temporal matching in highly dynamic and complex scenes.

\clearpage
{
    \small
    \bibliographystyle{ieeenat_fullname}
    \bibliography{main}

\begin{thebibliography}{75}
\providecommand{\natexlab}[1]{#1}
\providecommand{\url}[1]{\texttt{#1}}
\expandafter\ifx\csname urlstyle\endcsname\relax
  \providecommand{\doi}[1]{doi: #1}\else
  \providecommand{\doi}{doi: \begingroup \urlstyle{rm}\Url}\fi

\bibitem[Balntas et~al.(2017)Balntas, Lenc, Vedaldi, and
  Mikolajczyk]{balntas2017hpatches}
Vassileios Balntas, Karel Lenc, Andrea Vedaldi, and Krystian Mikolajczyk.
\newblock Hpatches: A benchmark and evaluation of handcrafted and learned local
  descriptors.
\newblock In \emph{Proceedings of the IEEE conference on computer vision and
  pattern recognition}, pages 5173--5182, 2017.

\bibitem[Bay et~al.(2008)Bay, Ess, Tuytelaars, and Van~Gool]{bay2008surf}
Herbert Bay, Andreas Ess, Tinne Tuytelaars, and Luc Van~Gool.
\newblock Speeded-up robust features (surf).
\newblock \emph{Computer vision and image understanding}, 110\penalty0
  (3):\penalty0 346--359, 2008.

\bibitem[Caron et~al.(2021)Caron, Touvron, Misra, J{\'e}gou, Mairal,
  Bojanowski, and Joulin]{caron2021dino}
Mathilde Caron, Hugo Touvron, Ishan Misra, Herv{\'e} J{\'e}gou, Julien Mairal,
  Piotr Bojanowski, and Armand Joulin.
\newblock Emerging properties in self-supervised vision transformers.
\newblock In \emph{Proceedings of the IEEE/CVF international conference on
  computer vision}, pages 9650--9660, 2021.

\bibitem[Chen et~al.(2022)Chen, Luo, Zhou, Tian, Zhen, Fang, Mckinnon, Tsin,
  and Quan]{chen2022aspanformer}
Hongkai Chen, Zixin Luo, Lei Zhou, Yurun Tian, Mingmin Zhen, Tian Fang, David
  Mckinnon, Yanghai Tsin, and Long Quan.
\newblock Aspanformer: Detector-free image matching with adaptive span
  transformer.
\newblock In \emph{European Conference on Computer Vision}, pages 20--36.
  Springer Nature Switzerland Cham, 2022.

\bibitem[Cho et~al.(2022)Cho, Hong, and Kim]{cho2022cats++}
Seokju Cho, Sunghwan Hong, and Seungryong Kim.
\newblock Cats++: Boosting cost aggregation with convolutions and transformers.
\newblock \emph{IEEE Transactions on Pattern Analysis and Machine
  Intelligence}, 45\penalty0 (6):\penalty0 7174--7194, 2022.

\bibitem[Cho et~al.(2024)Cho, Huang, Nam, An, Kim, and Lee]{cho2024locotrack}
Seokju Cho, Jiahui Huang, Jisu Nam, Honggyu An, Seungryong Kim, and Joon-Young
  Lee.
\newblock Local all-pair correspondence for point tracking.
\newblock \emph{arXiv preprint arXiv:2407.15420}, 2024.

\bibitem[Choy et~al.(2016)Choy, Gwak, Savarese, and Chandraker]{choy2016ucn}
Christopher~B Choy, JunYoung Gwak, Silvio Savarese, and Manmohan Chandraker.
\newblock Universal correspondence network.
\newblock \emph{Advances in neural information processing systems}, 29, 2016.

\bibitem[Chum et~al.(2003)Chum, Matas, and Kittler]{chum2003loransac}
Ond{\v{r}}ej Chum, Ji{\v{r}}{\'\i} Matas, and Josef Kittler.
\newblock Locally optimized ransac.
\newblock In \emph{Pattern Recognition: 25th DAGM Symposium, Magdeburg,
  Germany, September 10-12, 2003. Proceedings 25}, pages 236--243. Springer,
  2003.

\bibitem[Dai et~al.(2017)Dai, Chang, Savva, Halber, Funkhouser, and
  Nie{\ss}ner]{dai2017scannet}
Angela Dai, Angel~X Chang, Manolis Savva, Maciej Halber, Thomas Funkhouser, and
  Matthias Nie{\ss}ner.
\newblock Scannet: Richly-annotated 3d reconstructions of indoor scenes.
\newblock In \emph{Proceedings of the IEEE conference on computer vision and
  pattern recognition}, pages 5828--5839, 2017.

\bibitem[DeTone et~al.(2018)DeTone, Malisiewicz, and
  Rabinovich]{detone2018superpoint}
Daniel DeTone, Tomasz Malisiewicz, and Andrew Rabinovich.
\newblock Superpoint: Self-supervised interest point detection and description.
\newblock In \emph{Proceedings of the IEEE conference on computer vision and
  pattern recognition workshops}, pages 224--236, 2018.

\bibitem[Dhariwal and Nichol(2021)]{dhariwal2021amdiffusion}
Prafulla Dhariwal and Alexander Nichol.
\newblock Diffusion models beat gans on image synthesis.
\newblock \emph{Advances in neural information processing systems},
  34:\penalty0 8780--8794, 2021.

\bibitem[Doersch et~al.(2022)Doersch, Gupta, Markeeva, Recasens, Smaira, Aytar,
  Carreira, Zisserman, and Yang]{doersch2022tapvid}
Carl Doersch, Ankush Gupta, Larisa Markeeva, Adria Recasens, Lucas Smaira,
  Yusuf Aytar, Joao Carreira, Andrew Zisserman, and Yi Yang.
\newblock Tap-vid: A benchmark for tracking any point in a video.
\newblock \emph{Advances in Neural Information Processing Systems},
  35:\penalty0 13610--13626, 2022.

\bibitem[Dusmanu et~al.(2019)Dusmanu, Rocco, Pajdla, Pollefeys, Sivic, Torii,
  and Sattler]{dusmanu2019d2net}
Mihai Dusmanu, Ignacio Rocco, Tomas Pajdla, Marc Pollefeys, Josef Sivic,
  Akihiko Torii, and Torsten Sattler.
\newblock D2-net: A trainable cnn for joint description and detection of local
  features.
\newblock In \emph{Proceedings of the ieee/cvf conference on computer vision
  and pattern recognition}, pages 8092--8101, 2019.

\bibitem[Edstedt et~al.(2023)Edstedt, Athanasiadis, Wadenb{\"a}ck, and
  Felsberg]{edstedt2023dkm}
Johan Edstedt, Ioannis Athanasiadis, M{\aa}rten Wadenb{\"a}ck, and Michael
  Felsberg.
\newblock Dkm: Dense kernelized feature matching for geometry estimation.
\newblock In \emph{Proceedings of the IEEE/CVF Conference on Computer Vision
  and Pattern Recognition}, pages 17765--17775, 2023.

\bibitem[Edstedt et~al.(2024)Edstedt, Sun, B{\"o}kman, Wadenb{\"a}ck, and
  Felsberg]{edstedt2024roma}
Johan Edstedt, Qiyu Sun, Georg B{\"o}kman, M{\aa}rten Wadenb{\"a}ck, and
  Michael Felsberg.
\newblock Roma: Robust dense feature matching.
\newblock In \emph{Proceedings of the IEEE/CVF Conference on Computer Vision
  and Pattern Recognition}, pages 19790--19800, 2024.

\bibitem[Germain et~al.(2020)Germain, Bourmaud, and Lepetit]{germain2020s2dnet}
Hugo Germain, Guillaume Bourmaud, and Vincent Lepetit.
\newblock S2dnet: Learning accurate correspondences for sparse-to-dense feature
  matching.
\newblock \emph{arXiv preprint arXiv:2004.01673}, 2020.

\bibitem[Ham et~al.(2016)Ham, Cho, Schmid, and Ponce]{ham2016pfwillow}
Bumsub Ham, Minsu Cho, Cordelia Schmid, and Jean Ponce.
\newblock Proposal flow.
\newblock In \emph{Proceedings of the IEEE Conference on Computer Vision and
  Pattern Recognition}, pages 3475--3484, 2016.

\bibitem[Ham et~al.(2017)Ham, Cho, Schmid, and Ponce]{ham2017pfpascal}
Bumsub Ham, Minsu Cho, Cordelia Schmid, and Jean Ponce.
\newblock Proposal flow: Semantic correspondences from object proposals.
\newblock \emph{IEEE transactions on pattern analysis and machine
  intelligence}, 40\penalty0 (7):\penalty0 1711--1725, 2017.

\bibitem[Harley et~al.(2022)Harley, Fang, and Fragkiadaki]{harley2022particle}
Adam~W. Harley, Zhaoyuan Fang, and Katerina Fragkiadaki.
\newblock Particle video revisited: {T}racking through occlusions using point
  trajectories.
\newblock In \emph{ECCV}, 2022.

\bibitem[Hedlin et~al.(2024)Hedlin, Sharma, Mahajan, Isack, Kar, Tagliasacchi,
  and Yi]{hedlin2024usc}
Eric Hedlin, Gopal Sharma, Shweta Mahajan, Hossam Isack, Abhishek Kar, Andrea
  Tagliasacchi, and Kwang~Moo Yi.
\newblock Unsupervised semantic correspondence using stable diffusion.
\newblock \emph{Advances in Neural Information Processing Systems}, 36, 2024.

\bibitem[Huang et~al.(2022)Huang, Yang, He, Zhang, He, and
  Shrivastava]{huang2022scorrsan}
Shuaiyi Huang, Luyu Yang, Bo He, Songyang Zhang, Xuming He, and Abhinav
  Shrivastava.
\newblock Learning semantic correspondence with sparse annotations.
\newblock In \emph{European Conference on Computer Vision}, pages 267--284.
  Springer, 2022.

\bibitem[Ilharco et~al.(2021)Ilharco, Wortsman, Wightman, Gordon, Carlini,
  Taori, Dave, Shankar, Namkoong, Miller, et~al.]{ilharco2021openclip}
Gabriel Ilharco, Mitchell Wortsman, Ross Wightman, Cade Gordon, Nicholas
  Carlini, Rohan Taori, Achal Dave, Vaishaal Shankar, Hongseok Namkoong, John
  Miller, et~al.
\newblock Openclip, 2021.

\bibitem[Jiang et~al.(2024)Jiang, Karpur, Cao, Huang, and
  Araujo]{jiang2024omniglue}
Hanwen Jiang, Arjun Karpur, Bingyi Cao, Qixing Huang, and Andr{\'e} Araujo.
\newblock Omniglue: Generalizable feature matching with foundation model
  guidance.
\newblock In \emph{Proceedings of the IEEE/CVF Conference on Computer Vision
  and Pattern Recognition}, pages 19865--19875, 2024.

\bibitem[Karaev et~al.(2023)Karaev, Rocco, Graham, Neverova, Vedaldi, and
  Rupprecht]{karaev2023cotracker}
Nikita Karaev, Ignacio Rocco, Benjamin Graham, Natalia Neverova, Andrea
  Vedaldi, and Christian Rupprecht.
\newblock Cotracker: It is better to track together.
\newblock \emph{arXiv preprint arXiv:2307.07635}, 2023.

\bibitem[Kim et~al.(2013)Kim, Liu, Sha, and Grauman]{kim2013dsp}
Jaechul Kim, Ce Liu, Fei Sha, and Kristen Grauman.
\newblock Deformable spatial pyramid matching for fast dense correspondences.
\newblock In \emph{Proceedings of the IEEE Conference on Computer Vision and
  Pattern Recognition}, pages 2307--2314, 2013.

\bibitem[Kim et~al.(2017)Kim, Min, Ham, Jeon, Lin, and Sohn]{kim2017fcss}
Seungryong Kim, Dongbo Min, Bumsub Ham, Sangryul Jeon, Stephen Lin, and
  Kwanghoon Sohn.
\newblock Fcss: Fully convolutional self-similarity for dense semantic
  correspondence.
\newblock In \emph{Proceedings of the IEEE conference on computer vision and
  pattern recognition}, pages 6560--6569, 2017.

\bibitem[Kim et~al.(2019)Kim, Min, Jeong, Kim, Jeon, and Sohn]{kim2019semantic}
Seungryong Kim, Dongbo Min, Somi Jeong, Sunok Kim, Sangryul Jeon, and Kwanghoon
  Sohn.
\newblock Semantic attribute matching networks.
\newblock In \emph{Proceedings of the IEEE/CVF Conference on Computer Vision
  and Pattern Recognition}, pages 12339--12348, 2019.

\bibitem[Larsson(2020)]{larsson2020poselib}
Viktor Larsson.
\newblock Poselib-minimal solvers for camera pose estimation, 2020.

\bibitem[Lee et~al.(2019)Lee, Kim, Ponce, and Ham]{lee2019sfnet}
Junghyup Lee, Dohyung Kim, Jean Ponce, and Bumsub Ham.
\newblock Sfnet: Learning object-aware semantic correspondence.
\newblock In \emph{Proceedings of the IEEE/CVF Conference on Computer Vision
  and Pattern Recognition}, pages 2278--2287, 2019.

\bibitem[Leroy et~al.(2024)Leroy, Cabon, and Revaud]{leroy2024mast3r}
Vincent Leroy, Yohann Cabon, and J{\'e}r{\^o}me Revaud.
\newblock Grounding image matching in 3d with mast3r.
\newblock \emph{European Conference on Computer Vision}, 2024.

\bibitem[Li et~al.(2024)Li, Lu, Han, and Prisacariu]{li2024sd4match}
Xinghui Li, Jingyi Lu, Kai Han, and Victor~Adrian Prisacariu.
\newblock Sd4match: Learning to prompt stable diffusion model for semantic
  matching.
\newblock In \emph{Proceedings of the IEEE/CVF Conference on Computer Vision
  and Pattern Recognition}, pages 27558--27568, 2024.

\bibitem[Li and Snavely(2018)]{li2018megadepth}
Zhengqi Li and Noah Snavely.
\newblock Megadepth: Learning single-view depth prediction from internet
  photos.
\newblock In \emph{Proceedings of the IEEE conference on computer vision and
  pattern recognition}, pages 2041--2050, 2018.

\bibitem[Lindenberger et~al.(2023)Lindenberger, Sarlin, and
  Pollefeys]{lindenberger2023lightglue}
Philipp Lindenberger, Paul-Edouard Sarlin, and Marc Pollefeys.
\newblock Lightglue: Local feature matching at light speed.
\newblock In \emph{Proceedings of the IEEE/CVF International Conference on
  Computer Vision}, pages 17627--17638, 2023.

\bibitem[Liu et~al.(2010)Liu, Yuen, and Torralba]{liu2010siftflow}
Ce Liu, Jenny Yuen, and Antonio Torralba.
\newblock Sift flow: Dense correspondence across scenes and its applications.
\newblock \emph{IEEE transactions on pattern analysis and machine
  intelligence}, 33\penalty0 (5):\penalty0 978--994, 2010.

\bibitem[Loshchilov(2017)]{loshchilov2017adamw}
I Loshchilov.
\newblock Decoupled weight decay regularization.
\newblock \emph{arXiv preprint arXiv:1711.05101}, 2017.

\bibitem[Lowe(2004)]{lowe2004sift}
David~G Lowe.
\newblock Distinctive image features from scale-invariant keypoints.
\newblock \emph{International journal of computer vision}, 60:\penalty0
  91--110, 2004.

\bibitem[Luiten et~al.(2024)Luiten, Kopanas, Leibe, and
  Ramanan]{luiten2023dynamic}
Jonathon Luiten, Georgios Kopanas, Bastian Leibe, and Deva Ramanan.
\newblock Dynamic 3d gaussians: Tracking by persistent dynamic view synthesis.
\newblock In \emph{3DV}, 2024.

\bibitem[Luo et~al.(2024)Luo, Dunlap, Park, Holynski, and Darrell]{luo2024dhf}
Grace Luo, Lisa Dunlap, Dong~Huk Park, Aleksander Holynski, and Trevor Darrell.
\newblock Diffusion hyperfeatures: Searching through time and space for
  semantic correspondence.
\newblock \emph{Advances in Neural Information Processing Systems}, 36, 2024.

\bibitem[Luo et~al.(2019)Luo, Shen, Zhou, Zhang, Yao, Li, Fang, and
  Quan]{luo2019contextdesc}
Zixin Luo, Tianwei Shen, Lei Zhou, Jiahui Zhang, Yao Yao, Shiwei Li, Tian Fang,
  and Long Quan.
\newblock Contextdesc: Local descriptor augmentation with cross-modality
  context.
\newblock In \emph{Proceedings of the IEEE/CVF conference on computer vision
  and pattern recognition}, pages 2527--2536, 2019.

\bibitem[Min et~al.(2019)Min, Lee, Ponce, and Cho]{min2019spair}
Juhong Min, Jongmin Lee, Jean Ponce, and Minsu Cho.
\newblock Spair-71k: A large-scale benchmark for semantic correspondence.
\newblock \emph{arXiv preprint arXiv:1908.10543}, 2019.

\bibitem[Muja and Lowe(2014)]{muja2014nnm}
Marius Muja and David~G Lowe.
\newblock Scalable nearest neighbor algorithms for high dimensional data.
\newblock \emph{IEEE transactions on pattern analysis and machine
  intelligence}, 36\penalty0 (11):\penalty0 2227--2240, 2014.

\bibitem[Mur-Artal et~al.(2015)Mur-Artal, Montiel, and Tardos]{mur2015orbslam}
Raul Mur-Artal, Jose Maria~Martinez Montiel, and Juan~D Tardos.
\newblock Orb-slam: a versatile and accurate monocular slam system.
\newblock \emph{IEEE transactions on robotics}, 31\penalty0 (5):\penalty0
  1147--1163, 2015.

\bibitem[Ofri-Amar et~al.(2023)Ofri-Amar, Geyer, Kasten, and
  Dekel]{ofri2023neural}
Dolev Ofri-Amar, Michal Geyer, Yoni Kasten, and Tali Dekel.
\newblock Neural congealing: Aligning images to a joint semantic atlas.
\newblock In \emph{Proceedings of the IEEE/CVF Conference on Computer Vision
  and Pattern Recognition}, pages 19403--19412, 2023.

\bibitem[Oquab et~al.(2023)Oquab, Darcet, Moutakanni, Vo, Szafraniec, Khalidov,
  Fernandez, Haziza, Massa, El-Nouby, et~al.]{oquab2023dinov2}
Maxime Oquab, Timoth{\'e}e Darcet, Th{\'e}o Moutakanni, Huy Vo, Marc
  Szafraniec, Vasil Khalidov, Pierre Fernandez, Daniel Haziza, Francisco Massa,
  Alaaeldin El-Nouby, et~al.
\newblock Dinov2: Learning robust visual features without supervision.
\newblock \emph{arXiv preprint arXiv:2304.07193}, 2023.

\bibitem[Paszke et~al.(2019)Paszke, Gross, Massa, Lerer, Bradbury, Chanan,
  Killeen, Lin, Gimelshein, Antiga, et~al.]{paszke2019pytorch}
Adam Paszke, Sam Gross, Francisco Massa, Adam Lerer, James Bradbury, Gregory
  Chanan, Trevor Killeen, Zeming Lin, Natalia Gimelshein, Luca Antiga, et~al.
\newblock Pytorch: An imperative style, high-performance deep learning library.
\newblock \emph{Advances in neural information processing systems}, 32, 2019.

\bibitem[Potje et~al.(2024)Potje, Cadar, Araujo, Martins, and
  Nascimento]{potje2024xfeat}
Guilherme Potje, Felipe Cadar, Andr{\'e} Araujo, Renato Martins, and Erickson~R
  Nascimento.
\newblock Xfeat: Accelerated features for lightweight image matching.
\newblock In \emph{Proceedings of the IEEE/CVF Conference on Computer Vision
  and Pattern Recognition}, pages 2682--2691, 2024.

\bibitem[Radford et~al.(2021)Radford, Kim, Hallacy, Ramesh, Goh, Agarwal,
  Sastry, Askell, Mishkin, Clark, et~al.]{radford2021clip}
Alec Radford, Jong~Wook Kim, Chris Hallacy, Aditya Ramesh, Gabriel Goh,
  Sandhini Agarwal, Girish Sastry, Amanda Askell, Pamela Mishkin, Jack Clark,
  et~al.
\newblock Learning transferable visual models from natural language
  supervision.
\newblock In \emph{International conference on machine learning}, pages
  8748--8763. PMLR, 2021.

\bibitem[Revaud et~al.(2019)Revaud, De~Souza, Humenberger, and
  Weinzaepfel]{revaud2019r2d2}
Jerome Revaud, Cesar De~Souza, Martin Humenberger, and Philippe Weinzaepfel.
\newblock R2d2: Reliable and repeatable detector and descriptor.
\newblock \emph{Advances in neural information processing systems}, 32, 2019.

\bibitem[Rocco et~al.(2017)Rocco, Arandjelovic, and Sivic]{rocco2017geometric}
Ignacio Rocco, Relja Arandjelovic, and Josef Sivic.
\newblock Convolutional neural network architecture for geometric matching.
\newblock In \emph{Proceedings of the IEEE conference on computer vision and
  pattern recognition}, pages 6148--6157, 2017.

\bibitem[Rocco et~al.(2018{\natexlab{a}})Rocco, Arandjelovi{\'c}, and
  Sivic]{rocco2018geometric}
Ignacio Rocco, Relja Arandjelovi{\'c}, and Josef Sivic.
\newblock End-to-end weakly-supervised semantic alignment.
\newblock In \emph{Proceedings of the IEEE Conference on Computer Vision and
  Pattern Recognition}, pages 6917--6925, 2018{\natexlab{a}}.

\bibitem[Rocco et~al.(2018{\natexlab{b}})Rocco, Cimpoi, Arandjelovi{\'c},
  Torii, Pajdla, and Sivic]{rocco2018ncnet}
Ignacio Rocco, Mircea Cimpoi, Relja Arandjelovi{\'c}, Akihiko Torii, Tomas
  Pajdla, and Josef Sivic.
\newblock Neighbourhood consensus networks.
\newblock \emph{Advances in neural information processing systems}, 31,
  2018{\natexlab{b}}.

\bibitem[Rombach et~al.(2022{\natexlab{a}})Rombach, Blattmann, Lorenz, Esser,
  and Ommer]{rombach2022high}
Robin Rombach, Andreas Blattmann, Dominik Lorenz, Patrick Esser, and Bj{\"o}rn
  Ommer.
\newblock High-resolution image synthesis with latent diffusion models.
\newblock In \emph{Proceedings of the IEEE/CVF conference on computer vision
  and pattern recognition}, pages 10684--10695, 2022{\natexlab{a}}.

\bibitem[Rombach et~al.(2022{\natexlab{b}})Rombach, Blattmann, Lorenz, Esser,
  and Ommer]{rombach2022stablediffusion}
Robin Rombach, Andreas Blattmann, Dominik Lorenz, Patrick Esser, and Bj{\"o}rn
  Ommer.
\newblock High-resolution image synthesis with latent diffusion models.
\newblock In \emph{Proceedings of the IEEE/CVF conference on computer vision
  and pattern recognition}, pages 10684--10695, 2022{\natexlab{b}}.

\bibitem[Sand and Teller(2008)]{sand2008particle}
Peter Sand and Seth Teller.
\newblock Particle video: Long-range motion estimation using point
  trajectories.
\newblock \emph{International journal of computer vision}, 80:\penalty0 72--91,
  2008.

\bibitem[Sarlin et~al.(2020)Sarlin, DeTone, Malisiewicz, and
  Rabinovich]{sarlin2020superglue}
Paul-Edouard Sarlin, Daniel DeTone, Tomasz Malisiewicz, and Andrew Rabinovich.
\newblock Superglue: Learning feature matching with graph neural networks.
\newblock In \emph{Proceedings of the IEEE/CVF conference on computer vision
  and pattern recognition}, pages 4938--4947, 2020.

\bibitem[Sattler et~al.(2018)Sattler, Maddern, Toft, Torii, Hammarstrand,
  Stenborg, Safari, Okutomi, Pollefeys, Sivic, et~al.]{sattler2018benchmarking}
Torsten Sattler, Will Maddern, Carl Toft, Akihiko Torii, Lars Hammarstrand,
  Erik Stenborg, Daniel Safari, Masatoshi Okutomi, Marc Pollefeys, Josef Sivic,
  et~al.
\newblock Benchmarking 6dof outdoor visual localization in changing conditions.
\newblock In \emph{Proceedings of the IEEE conference on computer vision and
  pattern recognition}, pages 8601--8610, 2018.

\bibitem[Schonberger and Frahm(2016)]{schonberger2016sfm}
Johannes~L Schonberger and Jan-Michael Frahm.
\newblock Structure-from-motion revisited.
\newblock In \emph{Proceedings of the IEEE conference on computer vision and
  pattern recognition}, pages 4104--4113, 2016.

\bibitem[Seidenschwarz et~al.(2024)Seidenschwarz, Zhou, Duisterhof, Ramanan,
  and Leal-Taix{\'e}]{seidenschwarz2024dynomo}
Jenny Seidenschwarz, Qunjie Zhou, Bardienus Duisterhof, Deva Ramanan, and Laura
  Leal-Taix{\'e}.
\newblock Dynomo: Online point tracking by dynamic online monocular gaussian
  reconstruction.
\newblock \emph{arXiv preprint arXiv:2409.02104}, 2024.

\bibitem[Sun et~al.(2021)Sun, Shen, Wang, Bao, and Zhou]{sun2021loftr}
Jiaming Sun, Zehong Shen, Yuang Wang, Hujun Bao, and Xiaowei Zhou.
\newblock Loftr: Detector-free local feature matching with transformers.
\newblock In \emph{Proceedings of the IEEE/CVF conference on computer vision
  and pattern recognition}, pages 8922--8931, 2021.

\bibitem[Tang et~al.(2023)Tang, Jia, Wang, Phoo, and Hariharan]{tang2023dift}
Luming Tang, Menglin Jia, Qianqian Wang, Cheng~Perng Phoo, and Bharath
  Hariharan.
\newblock Emergent correspondence from image diffusion.
\newblock \emph{Advances in Neural Information Processing Systems},
  36:\penalty0 1363--1389, 2023.

\bibitem[Tian et~al.(2019)Tian, Yu, Fan, Wu, Heijnen, and
  Balntas]{tian2019sosnet}
Yurun Tian, Xin Yu, Bin Fan, Fuchao Wu, Huub Heijnen, and Vassileios Balntas.
\newblock Sosnet: Second order similarity regularization for local descriptor
  learning.
\newblock In \emph{Proceedings of the IEEE/CVF conference on computer vision
  and pattern recognition}, pages 11016--11025, 2019.

\bibitem[Truong et~al.(2020)Truong, Danelljan, and Timofte]{truong2020glu}
Prune Truong, Martin Danelljan, and Radu Timofte.
\newblock Glu-net: Global-local universal network for dense flow and
  correspondences.
\newblock In \emph{Proceedings of the IEEE/CVF conference on computer vision
  and pattern recognition}, pages 6258--6268, 2020.

\bibitem[Tyszkiewicz et~al.(2020)Tyszkiewicz, Fua, and
  Trulls]{tyszkiewicz2020disk}
Micha{\l} Tyszkiewicz, Pascal Fua, and Eduard Trulls.
\newblock Disk: Learning local features with policy gradient.
\newblock \emph{Advances in Neural Information Processing Systems},
  33:\penalty0 14254--14265, 2020.

\bibitem[Vaswani(2017)]{vaswani2017attention}
A Vaswani.
\newblock Attention is all you need.
\newblock \emph{Advances in Neural Information Processing Systems}, 2017.

\bibitem[Wang et~al.(2023)Wang, Chang, Cai, Li, Hariharan, Holynski, and
  Snavely]{wang2023omnimotion}
Qianqian Wang, Yen-Yu Chang, Ruojin Cai, Zhengqi Li, Bharath Hariharan,
  Aleksander Holynski, and Noah Snavely.
\newblock Tracking everything everywhere all at once.
\newblock In \emph{International Conference on Computer Vision}, 2023.

\bibitem[Wang et~al.(2024)Wang, Ye, Gao, Austin, Li, and
  Kanazawa]{som2024shapeofmotion}
Qianqian Wang, Vickie Ye, Hang Gao, Jake Austin, Zhengqi Li, and Angjoo
  Kanazawa.
\newblock Shape of motion: 4d reconstruction from a single video.
\newblock In \emph{arXiv preprint arXiv:2407.13764}, 2024.

\bibitem[Weinzaepfel et~al.(2022)Weinzaepfel, Leroy, Lucas, Br{\'e}gier, Cabon,
  Arora, Antsfeld, Chidlovskii, Csurka, and Revaud]{weinzaepfel2022croco}
Philippe Weinzaepfel, Vincent Leroy, Thomas Lucas, Romain Br{\'e}gier, Yohann
  Cabon, Vaibhav Arora, Leonid Antsfeld, Boris Chidlovskii, Gabriela Csurka,
  and J{\'e}r{\^o}me Revaud.
\newblock Croco: Self-supervised pre-training for 3d vision tasks by cross-view
  completion.
\newblock \emph{Advances in Neural Information Processing Systems},
  35:\penalty0 3502--3516, 2022.

\bibitem[Xiao et~al.(2024)Xiao, Wang, Zhang, Xue, Peng, Shen, and
  Zhou]{Xiao2024spatialtracker}
Yuxi Xiao, Qianqian Wang, Shangzhan Zhang, Nan Xue, Sida Peng, Yujun Shen, and
  Xiaowei Zhou.
\newblock Spatialtracker: Tracking any 2d pixels in 3d space.
\newblock In \emph{CVPR}, 2024.

\bibitem[Xu et~al.(2022)Xu, Lin, Zhang, Wang, and Li]{xu2022rnnpose}
Yan Xu, Kwan-Yee Lin, Guofeng Zhang, Xiaogang Wang, and Hongsheng Li.
\newblock Rnnpose: Recurrent 6-dof object pose refinement with robust
  correspondence field estimation and pose optimization.
\newblock In \emph{Proceedings of the IEEE/CVF conference on computer vision
  and pattern recognition}, pages 14880--14890, 2022.

\bibitem[Xue et~al.(2023)Xue, Budvytis, and Cipolla]{xue2023sfd2}
Fei Xue, Ignas Budvytis, and Roberto Cipolla.
\newblock Sfd2: Semantic-guided feature detection and description.
\newblock In \emph{Proceedings of the IEEE/CVF Conference on Computer Vision
  and Pattern Recognition}, pages 5206--5216, 2023.

\bibitem[Yi et~al.(2016)Yi, Trulls, Lepetit, and Fua]{yi2016lift}
Kwang~Moo Yi, Eduard Trulls, Vincent Lepetit, and Pascal Fua.
\newblock Lift: Learned invariant feature transform.
\newblock In \emph{Computer Vision--ECCV 2016: 14th European Conference,
  Amsterdam, The Netherlands, October 11-14, 2016, Proceedings, Part VI 14},
  pages 467--483. Springer International Publishing, 2016.

\bibitem[Yu et~al.(2021)Yu, Xu, Zhang, Zhao, Guan, and Tao]{yu2021ap}
Hang Yu, Yufei Xu, Jing Zhang, Wei Zhao, Ziyu Guan, and Dacheng Tao.
\newblock Ap-10k: A benchmark for animal pose estimation in the wild.
\newblock \emph{arXiv preprint arXiv:2108.12617}, 2021.

\bibitem[Zhang et~al.(2024{\natexlab{a}})Zhang, Herrmann, Hur, Chen, Jampani,
  Sun, and Yang]{zhang2024geosddino}
Junyi Zhang, Charles Herrmann, Junhwa Hur, Eric Chen, Varun Jampani, Deqing
  Sun, and Ming-Hsuan Yang.
\newblock Telling left from right: Identifying geometry-aware semantic
  correspondence.
\newblock In \emph{Proceedings of the IEEE/CVF Conference on Computer Vision
  and Pattern Recognition}, pages 3076--3085, 2024{\natexlab{a}}.

\bibitem[Zhang et~al.(2024{\natexlab{b}})Zhang, Herrmann, Hur, Polania~Cabrera,
  Jampani, Sun, and Yang]{zhang2024sddino}
Junyi Zhang, Charles Herrmann, Junhwa Hur, Luisa Polania~Cabrera, Varun
  Jampani, Deqing Sun, and Ming-Hsuan Yang.
\newblock A tale of two features: Stable diffusion complements dino for
  zero-shot semantic correspondence.
\newblock \emph{Advances in Neural Information Processing Systems}, 36,
  2024{\natexlab{b}}.

\bibitem[Zhou et~al.(2021)Zhou, Sattler, and Leal-Taixe]{zhou2021patch2pix}
Qunjie Zhou, Torsten Sattler, and Laura Leal-Taixe.
\newblock Patch2pix: Epipolar-guided pixel-level correspondences.
\newblock In \emph{Proceedings of the IEEE/CVF conference on computer vision
  and pattern recognition}, pages 4669--4678, 2021.

\end{thebibliography}
}

\end{document}